\documentclass[10pt, a4paper, logo]{googledeepmind}
\usepackage{times}
\usepackage{subcaption}
\usepackage{pgfplots}
\pgfplotsset{compat=1.18}
\usepackage[most]{tcolorbox}
\tcbuselibrary{breakable}
\usepackage{natbib}
\usepackage[capitalise]{cleveref}
\crefname{figure}{Fig.}{Figs.}
\crefname{table}{Table}{Tables}
\crefname{appendix}{App.}{App.}
\crefname{section}{§}{§§}
\crefname{equation}{Eq.}{Eqs.}
\usepackage{array}
\usepackage{multirow}
\usepackage{stfloats}
\usepackage{float}
\usepackage{etoolbox}
\usepackage[normalem]{ulem}
\usepackage{makecell}
\usepackage{wrapfig}
\usepackage{placeins}
\usepackage{soul,xcolor}

\usepackage{amsmath,amsfonts,bm}
\usepackage{natbib}
\usepackage{fancyhdr}
\def\eqref#1{equation~\ref{#1}}
\def\1{\bm{1}}

\DeclareMathAlphabet{\mathsfit}{\encodingdefault}{\sfdefault}{m}{sl}
\SetMathAlphabet{\mathsfit}{bold}{\encodingdefault}{\sfdefault}{bx}{n}

\newcommand\myparagraph[1]{
\vskip 0.05in
\noindent{\bf {#1}}}

\definecolor{ArrowGreen}{HTML}{2E7D32}
\definecolor{ArrowRed}{HTML}{C62828}
\definecolor{Green3}{RGB}{0,205,0}
\definecolor{Red1}{RGB}{255,0,0}

\newcommand{\systemname}{Trace2Skill}
\title{\systemname: Distill Trajectory-Local Lessons into Transferable Agent Skills}

\author[2,3]{Jingwei Ni\textsuperscript{$\S$,*}}
\author[4]{Yihao Liu\textsuperscript{$\S$,*}}
\author[4]{Xinpeng Liu\textsuperscript{$\S$,*}}
\author[5]{Yutao Sun\textsuperscript{$\S$,*}}
\author[1]{Mengyu Zhou\textsuperscript{$\dag$}}
\author[1]{Pengyu Cheng}
\author[1]{Dexin Wang}
\author[1]{Erchao Zhao}
\author[1]{Xiaoxi Jiang}
\author[1]{Guanjun Jiang}
\affil[1]{Qwen Large Model Application Team, Alibaba}
\affil[2]{ETH Zürich}
\affil[3]{University of Zurich}
\affil[4]{Peking University}
\affil[5]{Zhejiang University}
\footnotetext{\textsuperscript{$\S$}Core Contributors.}  % 同等贡献脚注

\correspondingauthor{zhoumengyu.zmy@alibaba-inc.com}

\begin{abstract}
Large Language Model (LLM) agents increasingly rely on domain-specific skills, yet manually authoring such skills does not scale, and skills generated purely from parametric knowledge often miss critical operational pitfalls. We introduce \systemname{}, a framework that consolidates broad execution trajectories in parallel into a unified skill directory through inductive reasoning over agent experience. \systemname{} supports both deepening existing human-written skills and creating useful skills from weak LLM-generated drafts. Experiments demonstrate the effectiveness of \systemname{} across diverse domains, including office workflows, math reasoning, and vision QA. Importantly, the evolved skills are not merely memorized artifacts of the trajectories used to create them: they often transfer across model scales, across model families, and to out-of-distribution settings. For example, skills evolved from Qwen3.5-35B trajectories improve a Qwen3.5-122B agent by up to $57.65$ percentage points on WikiTableQuestions. Further analyses show that \systemname{} outperforms sequential skill editing and ReasoningBank-style retrieval memories, compresses recurring failures and workarounds into standard operating procedures (SoPs), and yields portable skills that can be reused without parameter updates or test-time retrieval.\footnote{Code: \url{https://github.com/Qwen-Applications/Trace2Skill}}

\end{abstract}

\begin{document}
\maketitle

% intro_v2.tex — Failure-Grounded Distillation framing
% Central question: distill vs. retrieve. Asymmetry finding moved up. No "first X" claims.
\section{Introduction}
\label{sec:introduction}

\begin{wrapfigure}{r}{0.5\linewidth}
\centering
\includegraphics[width=\linewidth]{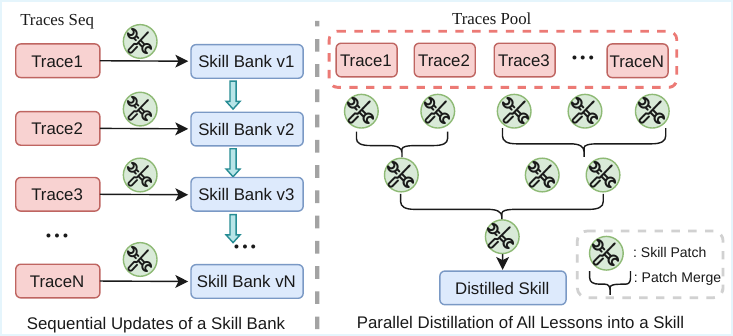}
\caption{\textbf{Left:} sequential online skill evolution edits the skill after each incoming trace. \textbf{Right:} \systemname{} analyzes many traces in parallel and hierarchically consolidates recurring lessons into one portable skill.}
\label{fig:intro}
\end{wrapfigure}

LLM-based agents increasingly rely on \emph{skills}: reusable documents that encode task procedures, domain knowledge, and operational guidelines \citep{anthropic2026skills,zhou2026comprehensivesurveyagentskills}.
As agents move into specialized file- and tool-use workflows, the bottleneck is no longer only model capability, but also the ability to create and maintain high-quality skills for each domain \citep{han2026sweskillsbenchagentskillsactually,li2026organizingorchestratingbenchmarkingagent,anthropic2026skillcreatorconversation,liang2026skillnetcreateevaluateconnect,zhou2026comprehensivesurveyagentskills}.
Human-written skills can help, but they are not uniformly beneficial across agents: in \cref{tab:main_v1}, the official \texttt{xlsx} skill improves a 122B spreadsheet agent while hurting a 35B agent on the same benchmark.
Generating skills purely from parametric knowledge is also brittle, because such drafts often lack the concrete failure modes, workarounds, and operational details exposed by actual execution traces \citep{li2026skillsbenchbenchmarkingagentskills,jiang2026xskillcontinuallearningexperience,zhou2026comprehensivesurveyagentskills}.

Recent systems therefore use agent experience to evolve skills or memories online \citep{yang2026autoskillexperiencedrivenlifelonglearning,xia2026skillrlevolvingagentsrecursive,alzubi2026evoskillautomatedskilldiscovery,zhou2026mementoskillsletagentsdesign,jiang2026xskillcontinuallearningexperience,zhou2026comprehensivesurveyagentskills}.
This direction is promising, but many existing approaches either store trajectory-local lessons for retrieval or edit skills sequentially as new traces arrive (\cref{fig:intro}, left).
Such designs can fragment reusable knowledge across a large memory or skill collection, and sequential editing can make later skills depend on the order of earlier updates \citep{li2026singleagentskillsreplacemultiagent}.
\textbf{Human experts usually work differently:} they inspect broad traces, abstract recurring patterns, and write compact procedures that are reusable across cases.

We introduce \systemname{}, a framework for turning execution traces into portable skills.
Rather than retrieving per-episode memories at test time or absorbing traces through order-dependent sequential edits, \systemname{} analyzes many traces jointly and consolidates recurring lessons into a single skill directory (\cref{fig:intro}, right).
The same mechanism supports two common use cases: \emph{deepening} an existing human-written skill and \emph{creating} a useful skill from a weak LLM-generated draft.
This many-to-one consolidation acts as an inductive reasoning step over agent experience \citep{xiong-etal-2025-co,li2025mirageevaluatingexplaininginductive,lin2025llmbasedscientificinductivereasoning}, while preserving the portability of ordinary skill files.

Experiments show that trajectory-grounded skills improve performance while remaining portable across models, benchmarks, and task domains.
In spreadsheet workflows, \systemname{} both strengthens Anthropic's official \texttt{xlsx} skill and creates useful skills from scratch.
The resulting skills transfer across model scales and families: for example, a skill evolved with Gemma-4-31B-it \citep{gemma42026} improves Qwen3.5-122B \citep{qwen35blog}.
They also generalize to out-of-distribution (OOD) data and tasks, such as transferring from spreadsheet editing to table QA.
The benefits extend beyond spreadsheets: \systemname{} improves math reasoning and DocVQA, and further strengthens Anthropic's official PDF, DOCX, and PPTX workflow skills.
Further analyses show that:
parallel consolidation is much faster and generally stronger than order-dependent skill editing; a single distilled skill outperforms ReasoningBank-style episodic retrieval \citep{ouyang2026reasoningbankscalingagentselfevolving}; and agentic analysts produce better patches by inspecting artifacts and validating fixes.
Qualitatively, the learned patches are not trajectory-specific tips: they coalesce into reusable SoPs, while patch-selection studies show that their value is often combinatorial, making holistic consolidation more reliable than greedy local selection. 

We contribute: (1) \textbf{Trace2Skill}, a framework for automatic skill creation and deepening. It mirrors human skill writing: building broad prior knowledge through extensive trajectory analysis before drafting skills (\cref{sec:method}). (2) \textbf{Broad empirical evidence} that trajectory-grounded evolution yields skills that transfer effectively across LLM scales, families, and OOD tasks (\cref{sec:experiments}). (3) \textbf{Comprehensive analysis} showing why consolidation works: parallel merging improves over sequential updates, one consolidated skill beats retrieval-based reasoning memories, agentic diagnosis improves patch quality, and patch value is often combinatorial (\cref{sec:analysis}).

\begin{figure}[t]
    \centering
    \includegraphics[width=\linewidth]{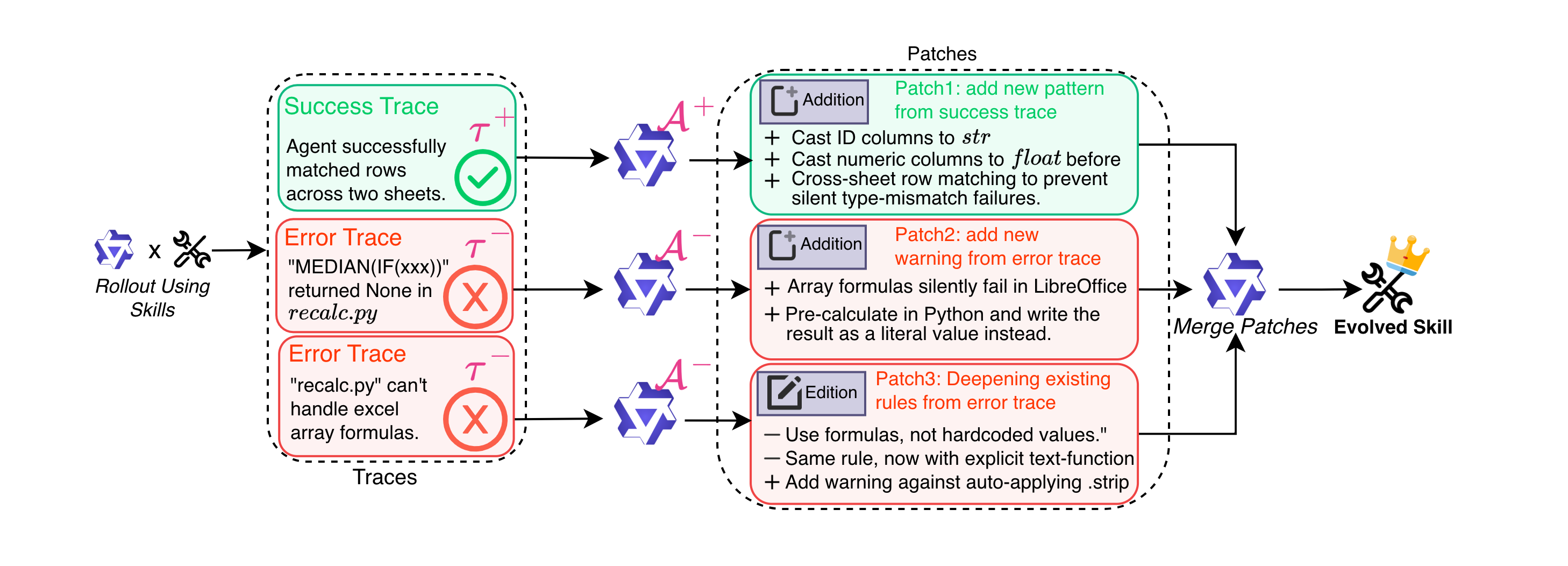}
    \caption{Overview of \systemname{}'s three-stage pipeline: (1)~roll out a frozen agent to collect labeled success and failure trajectories, (2)~propose trajectory-level skill patches in parallel with separate error and success analysts, and (3)~hierarchically merge all patches into one portable skill directory.}
    \label{fig:pipeline}
\end{figure}

% \begin{itemize}[leftmargin=*]
%     \item \textbf{Trace2Skill}, a framework for automatic skill creation and deepening. It mirrors human skill writing: building broad prior knowledge through extensive trajectory analysis before drafting skills (\cref{sec:method}).

%     \item \textbf{Broad empirical evidence} that trajectory-grounded evolution yields skills that transfer effectively across LLM scales, families, and out-of-distribution tasks (\cref{sec:experiments}).

%     \item \textbf{Comprehensive analysis} showing why consolidation works: parallel merging improves over sequential updates, one consolidated skill beats retrieval-based reasoning memories, agentic diagnosis improves patch quality, and patch value is often combinatorial (\cref{sec:analysis}).
% \end{itemize}

\section{\systemname}
\label{sec:method}

\cref{fig:pipeline} shows the three-stage \systemname{} pipeline: collect trajectories, propose trajectory-level patches in parallel, and consolidate them into one portable skill.
We first define the skill-evolution objective, then describe each stage.

\subsection{Skill and Problem Formalization}
\label{sec:method:skill}

A skill is a human-readable directory $\mathcal{S}=(M,\mathcal{R})$, where $M$ is the root \texttt{SKILL.md} and $\mathcal{R}$ contains auxiliary references, scripts, or assets.
The root document stores broadly applicable procedural knowledge, while auxiliary files provide deterministic tools or lower-frequency details.
Let $\pi_\theta$ be a fixed LLM agent using skill $\mathcal{S}$ at inference time, and let $\mathcal{P}(\mathcal{S};\pi_\theta,\mathcal{D})$ denote the pass rate of that agent on task set $\mathcal{D}$.
Given an evolving set $\mathcal{D}_\text{evolve}$ and a disjoint test set $\mathcal{D}_\text{test}$, skill evolution constructs a new skill from evolving-set trajectories without updating $\theta$:
\begin{equation}
    \mathcal{S}^* = \mathcal{E}(\mathcal{S}_0,\mathcal{D}_\text{evolve};\pi_\theta),
    \qquad
    \mathcal{P}(\mathcal{S}^*;\pi_\theta,\mathcal{D}_\text{test})
    > \mathcal{P}(\mathcal{S}_0;\pi_\theta,\mathcal{D}_\text{test}).
\end{equation}
We evaluate two initializations for $\mathcal{S}_0$: a human-expert skill and an LLM-generated draft from parametric knowledge alone.

\subsection{Stage 1: Trajectory Generation}
\label{sec:method:trajectories}

We use a ReAct-style harness \citep{yao2023reactsynergizingreasoningacting}.
For each evolving-set task, the fixed agent runs with $\mathcal{S}_0$ and produces a trajectory $\tau_i$ containing the query, reasoning/tool-use history, final output, and a binary correctness outcome.
The resulting corpus $\mathcal{T}$ is split into failures $\mathcal{T}^-$ and successes $\mathcal{T}^+$.
Trajectory generation is parallel across evolving-set problems, so the trace corpus can be collected independently before patch proposal.
See prompt templates in \cref{app:stage1_prompt}.

\subsection{Stage 2: Parallel Patch Proposal}
\label{sec:method:swarm}

A group of analyst sub-agents independently proposes skill patches from individual trajectories.
Failures are sent to an error analyst $\mathcal{A}^-$, successes to a success analyst $\mathcal{A}^+$, and the analysts read $\mathcal{S}_0$ and propose a patch for it, leading to a patch pool $\mathcal{P}=\mathcal{P}^-\cup\mathcal{P}^+$.

The analyst roles are intentionally asymmetric.
$\mathcal{A}^+$ uses a single-pass workflow to identify reusable behavior patterns from successful trajectories.
$\mathcal{A}^-$ uses a ReAct-style loop that can inspect traces and artifacts, compare outputs against ground truth, and validate candidate fixes before proposing a patch.
Failures that cannot be causally explained are excluded, ensuring $\mathcal{P}^-$ is grounded in verified failure mechanisms rather than log-only guesses \citep{ouyang2026reasoningbankscalingagentselfevolving}.
Both roles are prompted to write concise, actionable patches following skill-writing guidance \citep{anthropic2026skillcreatorconversation}.
Analyst prompt templates and representative example patches are provided in \cref{app:stage2_prompts}.

\subsection{Stage 3: Patch Consolidation}
\label{sec:method:consolidation}

Stage~3 consolidates the patch pool into one coherent update $p^*$ and applies it to $\mathcal{S}_0$.
Patches are merged hierarchically for $L=\lceil \log_{B_\text{merge}}|\mathcal{P}| \rceil$ levels; at each level $\ell$, up to $B_\text{merge}$ patches are synthesized into one patch:
\begin{equation}
    p^{(\ell+1)}
    = \mathcal{M}\!\left(\pi_\theta,\; \mathcal{S}_0,\; \{p_1^{(\ell)}, \ldots, p_{B_\text{merge}}^{(\ell)}\}\right),
    \qquad \ell=0,\ldots,L-1,
\end{equation}
where $\mathcal{M}$ deduplicates, resolves conflicts, and preserves non-overlapping insights.
$\pi_\theta$ itself serves as trajectory generator, analyst, and merge operator, so no external evolution/teacher model is required.
The final $p^*$ is translated into diff-style edits and applied with deterministic guardrails: reject edits to missing files, withhold line-range conflicts, and validate the updated skill format.

The merge also performs inductive generalization.
Because each patch comes from one trajectory, recurring edits across independent patches are evidence of systematic task properties rather than one-off fixes.
$\mathcal{M}$ is therefore instructed to prefer prevalent patterns and discard idiosyncratic edits, producing a compact skill that replaces $\mathcal{S}_0$ and is used directly at inference without retrieval. The merge operator prompt template and an example consolidated patch $p^*$ are given in \cref{app:stage3_prompts}.

% \subsection{Two Evolution Modes}
% \label{sec:method:modes}

\systemname{} supports two modes that cover scenarios with and without expert-written skills: \emph{Skill deepening} starts with a human-written skill, and \emph{Skill creation} starts from a skill generated from LLM parametric knowledge. Both improve the target skill using patches induced from trace analysis.

\begin{table*}[t]
\centering
\setlength{\tabcolsep}{3.5pt}
\caption{Main spreadsheet results across skill-author (evolves the skill) and skill-user (runs it at inference) models. SpreadsheetBench reports Vrf (verified), Soft (problem pass rate), and Hard (task pass rate); WikiTQ and HiTab are out-of-distribution (OOD) table-QA transfer, and $\mathrm{Avg}{=}(\mathrm{ID}_{\mathrm{avg}}{+}\mathrm{OOD}_{\mathrm{avg}})/2$ equally weights in-distribution (ID, SpreadsheetBench) and OOD. Reference rows are absolute scores; evolved rows are signed deltas from the Human-Written (Deepening) or Parametric (Creation) baseline. \textbf{Bold} marks the largest absolute score per column. Per-seed standard deviations are in \cref{tab:main_std}.}
\label{tab:main_v1}
\resizebox{\textwidth}{!}{%
\begin{tabular}{@{}l ccccc ccccc c@{}}
\toprule
& \multicolumn{5}{c}{\textit{Skill User: Qwen3.5-122B-A10B}}
& \multicolumn{5}{c}{\textit{Skill User: Qwen3.5-35B-A3B}}
& \\
\cmidrule(lr){2-6}\cmidrule(lr){7-11}
& \multicolumn{3}{c}{\textit{SpreadsheetBench}} & \multicolumn{2}{c}{\textit{OOD}}
& \multicolumn{3}{c}{\textit{SpreadsheetBench}} & \multicolumn{2}{c}{\textit{OOD}}
& \\
\cmidrule(lr){2-4}\cmidrule(lr){5-6}\cmidrule(lr){7-9}\cmidrule(lr){10-11}
\textbf{Condition}
    & \textbf{Vrf}$\uparrow$ & \textbf{Soft}$\uparrow$ & \textbf{Hard}$\uparrow$ & \textbf{WikiTQ}$\uparrow$ & \textbf{HiTab}$\uparrow$
    & \textbf{Vrf}$\uparrow$ & \textbf{Soft}$\uparrow$ & \textbf{Hard}$\uparrow$ & \textbf{WikiTQ}$\uparrow$ & \textbf{HiTab}$\uparrow$
    & \textbf{Avg}$\uparrow$ \\
\midrule
\multicolumn{12}{l}{\textit{Baseline (absolute scores)}} \\
\quad No Skill
    & 27.67 & 28.90 & 17.57 & 21.50 & 14.42 & 19.00 & 18.00 & 4.60 & 13.33 & 19.12 & 18.19 \\
\quad Human-Written
    & 48.33 & 36.30 & 17.03 & 74.68 & 41.31 & 9.67 & 13.03 & 3.37 & 9.02 & 5.30 & 26.93 \\
\quad Parametric
    & 26.17 & 36.60 & 17.50 & 23.73 & 17.36 & 20.17 & 13.70 & 3.87 & 20.14 & 13.94 & 19.23 \\
\midrule
\multicolumn{12}{l}{\textit{Skill Author: Qwen3.5-122B-A10B}} \\[2pt]
\multicolumn{12}{l}{\quad\textit{Deepening (Delta from init: Human-Written)}} \\
\quad\quad +Error
    & \cellcolor{Green3!46}+17.50 & \cellcolor{Green3!57}+10.30 & \cellcolor{Green3!50}+10.40 & \cellcolor{Green3!2}+1.62 & \cellcolor{Green3!5}+2.14 & \cellcolor{Green3!60}\textbf{+27.00} & \cellcolor{Green3!39}+9.44 & \cellcolor{Green3!24}+2.86 & \cellcolor{Green3!13}+9.26 & \cellcolor{Green3!16}+9.55 & \cellcolor{Green3!30}+9.28 \\
\quad\quad +Success
    & \cellcolor{Green3!47}+18.00 & \cellcolor{Green3!47}+8.60 & \cellcolor{Green3!42}+8.70 & \cellcolor{Red1!11}-10.35 & \cellcolor{Red1!2}-0.95 & \cellcolor{Green3!44}+19.66 & \cellcolor{Green3!28}+6.84 & \cellcolor{Green3!11}+1.33 & \cellcolor{Green3!17}+12.09 & \cellcolor{Green3!37}+22.33 & \cellcolor{Green3!26}+8.15 \\
\quad\quad +Combined
    & \cellcolor{Green3!57}\textbf{+21.50} & \cellcolor{Green3!60}\textbf{+10.87} & \cellcolor{Green3!60}\textbf{+12.50} & \cellcolor{Green3!5}+4.56 & \cellcolor{Green3!6}+2.97 & \cellcolor{Green3!47}+21.16 & \cellcolor{Green3!37}+8.84 & \cellcolor{Green3!15}+1.80 & \cellcolor{Green3!9}+6.64 & \cellcolor{Green3!20}+11.94 & \cellcolor{Green3!31}+9.65 \\[2pt]
\multicolumn{12}{l}{\quad\textit{Creation (Delta from init: Parametric)}} \\
\quad\quad +Error
    & \cellcolor{Green3!60}+22.83 & \cellcolor{Green3!21}+3.77 & \cellcolor{Green3!28}+5.87 & \cellcolor{Green3!8}+7.89 & \cellcolor{Red1!4}-1.70 & \cellcolor{Green3!19}+8.66 & \cellcolor{Green3!39}+9.53 & \cellcolor{Green3!33}+4.00 & \cellcolor{Green3!3}+2.06 & \cellcolor{Green3!16}+9.64 & \cellcolor{Green3!22}+6.79 \\
\quad\quad +Success
    & \cellcolor{Green3!40}+15.33 & \cellcolor{Red1!5}-0.93 & \cellcolor{Green3!21}+4.33 & \cellcolor{Green3!25}+23.70 & \cellcolor{Green3!42}+19.89 & \cellcolor{Green3!29}+12.83 & \cellcolor{Green3!48}+11.57 & \cellcolor{Green3!51}+6.13 & \cellcolor{Green3!42}+30.36 & \cellcolor{Green3!48}\textbf{+28.90} & \cellcolor{Green3!55}+16.96 \\
\quad\quad +Combined
    & \cellcolor{Green3!37}+14.00 & \cellcolor{Red1!3}-0.63 & \cellcolor{Green3!17}+3.53 & \cellcolor{Green3!34}+32.32 & \cellcolor{Green3!49}+23.11 & \cellcolor{Green3!34}+15.50 & \cellcolor{Green3!60}\textbf{+14.50} & \cellcolor{Green3!60}\textbf{+7.23} & \cellcolor{Green3!41}+29.70 & \cellcolor{Green3!46}+27.71 & \cellcolor{Green3!60}+18.62 \\[2pt]
\midrule
\multicolumn{12}{l}{\textit{Skill Author: Qwen3.5-35B-A3B}} \\[2pt]
\multicolumn{12}{l}{\quad\textit{Deepening (Delta from init: Human-Written)}} \\
\quad\quad +Error
    & \cellcolor{Green3!44}+16.67 & \cellcolor{Green3!47}+8.50 & \cellcolor{Green3!39}+8.14 & \cellcolor{Red1!7}-6.36 & \cellcolor{Red1!5}-2.38 & \cellcolor{Green3!39}+17.33 & \cellcolor{Green3!38}+9.17 & \cellcolor{Green3!40}+4.83 & \cellcolor{Green3!4}+2.71 & \cellcolor{Green3!13}+8.08 & \cellcolor{Green3!18}+5.64 \\
\quad\quad +Success
    & \cellcolor{Green3!6}+2.17 & \cellcolor{Green3!15}+2.73 & \cellcolor{Green3!16}+3.30 & \cellcolor{Green3!2}+1.46 & \cellcolor{Green3!4}+1.70 & \cellcolor{Green3!27}+12.33 & \cellcolor{Green3!24}+5.87 & \cellcolor{Green3!10}+1.23 & \cellcolor{Green3!60}\textbf{+43.23} & \cellcolor{Green3!57}+34.70 & \cellcolor{Green3!40}+12.44 \\
\quad\quad +Combined
    & \cellcolor{Green3!18}+6.67 & \cellcolor{Green3!21}+3.87 & \cellcolor{Green3!20}+4.17 & \cellcolor{Green3!3}+2.65 & \cellcolor{Green3!5}+2.44 & \cellcolor{Green3!44}+20.00 & \cellcolor{Green3!24}+5.77 & \cellcolor{Green3!20}+2.36 & \cellcolor{Green3!59}+42.20 & \cellcolor{Green3!60}+36.46 & \cellcolor{Green3!45}\textbf{+14.04} \\[2pt]
\multicolumn{12}{l}{\quad\textit{Creation (Delta from init: Parametric)}} \\
\quad\quad +Error
    & \cellcolor{Green3!3}+1.00 & \cellcolor{Red1!43}-7.70 & \cellcolor{Green3!5}+1.03 & \cellcolor{Green3!60}\textbf{+57.65} & \cellcolor{Green3!60}\textbf{+28.25} & \cellcolor{Green3!9}+3.83 & \cellcolor{Green3!30}+7.30 & \cellcolor{Green3!22}+2.66 & \cellcolor{Green3!18}+12.66 & \cellcolor{Green3!29}+17.81 & \cellcolor{Green3!49}+15.22 \\
\quad\quad +Success
    & \cellcolor{Green3!14}+5.33 & \cellcolor{Red1!25}-4.57 & \cellcolor{Green3!12}+2.43 & \cellcolor{Green3!9}+9.09 & \cellcolor{Green3!5}+2.34 & \cellcolor{Green3!13}+5.66 & \cellcolor{Green3!24}+5.80 & \cellcolor{Green3!22}+2.63 & \cellcolor{Green3!5}+3.31 & \cellcolor{Green3!31}+18.94 & \cellcolor{Green3!18}+5.65 \\
\quad\quad +Combined
    & \cellcolor{Green3!22}+8.33 & \cellcolor{Red1!32}-5.83 & \cellcolor{Green3!10}+2.00 & \cellcolor{Green3!32}+30.82 & \cellcolor{Green3!36}+16.93 & \cellcolor{Green3!10}+4.33 & \cellcolor{Green3!40}+9.73 & \cellcolor{Green3!39}+4.73 & \cellcolor{Green3!25}+18.00 & \cellcolor{Green3!42}+25.22 & \cellcolor{Green3!43}+13.31 \\[2pt]
\bottomrule
\end{tabular}
}%
\end{table*}

\section{Experiments}
\label{sec:experiments}

\subsection{Experimental Setup}
\label{sec:experiments:setup}

\myparagraph{Spreadsheet setup.}
Our main experiments use SpreadsheetBench-Verified \citep{ma2024spreadsheetbenchchallengingrealworld}, where agents manipulate \texttt{xlsx} files through tool use.
We split its 400 samples into 200 evolution problems and 200 held-out test problems, and also report Soft/Hard scores on full SpreadsheetBench (excluding all evolving-set problems) plus OOD transfer to WikiTableQuestions \citep{pasupat2015compositionalsemanticparsingsemistructured} and HiTab \citep{cheng-etal-2022-hitab} converted into spreadsheet format.
All spreadsheet results are averaged over three random seeds using each benchmark's official evaluation criteria.

\myparagraph{Skill settings.}
We compare \textbf{No Skill}, the Anthropic \texttt{xlsx} skill (\textbf{Human-Written}), LLM-generated skills using \textbf{Parametric} knowledge, and three \systemname{} variants: \textbf{+Error}, \textbf{+Success}, and \textbf{+Combined}, which consolidate patches from failed trajectories only, from successful trajectories only, and from all trajectories, respectively.
% \textbf{+Error} consolidates patches from failed trajectories, \textbf{+Success} consolidates reusable behaviors from successful trajectories, and \textbf{+Combined} merges both evidence sources.
\textit{Skill Deepening} starts from Human-Written, while \textit{Skill Creation} starts from Parametric.
We evaluate Qwen3.5-122B-A10B and Qwen3.5-35B-A3B as both skill authors and skill users. We do 100\% self-evolution: the same model generates trajectories, proposes patches, and edits skills.
Details of dataset construction, scoring, and model serving are in \cref{app:experimental_details}; the external \texttt{skill-creator} baseline is in \cref{app:skill_creator_baseline}.

\subsection{Main Results}
\label{sec:experiments:main}

\cref{tab:main_v1} reports all spreadsheet results across skill conditions, author models, user models, and transfer tasks.
Baseline rows give absolute scores; evolved rows give signed deltas from the relevant baseline, with Deepening compared to Human-Written and Creation compared to Parametric.
We use \textbf{Avg} as the primary summary metric because it equally weights ID and OOD performance across both user models, rewarding skills that generalize across models and tasks.

\myparagraph{Baselines reveal why both settings matter.}
Human-Written is a useful handcrafted prior, but it is not reliably portable across model scales; Parametric remains close to No Skill, confirming that parametric knowledge alone does not yield actionable spreadsheet skills \citep{han2026sweskillsbenchagentskillsactually}.
These baselines motivate the two evaluation regimes: Deepening tests whether a strong manual skill can be made more transferable, while Creation tests whether trajectory-grounded distillation can build a useful skill from a weak seed.

\myparagraph{Both Deepening and Creation produce generalizable skills.}
Starting from Human-Written, evolved skills consistently strengthen in-distribution spreadsheet performance and often transfer to other model scales and OOD table tasks; starting from Parametric, Creation can match or exceed Human-Written quality in favorable settings.
The gains are broad: they appear across author models, user models, and task families, not only on the model that produced the traces.
Notably, 35B Deepening +Combined attains the best absolute Avg, showing that skills authored by a small LLM can also generalize.
Across analysts, both +Error and +Success help spreadsheet tasks, and +Combined usually gives the largest Avg improvement.

\subsection{Model-Family Generalization}
\label{sec:experiments:model_family}

\begin{figure}[t]
    \centering
    \includegraphics[width=0.96\linewidth]{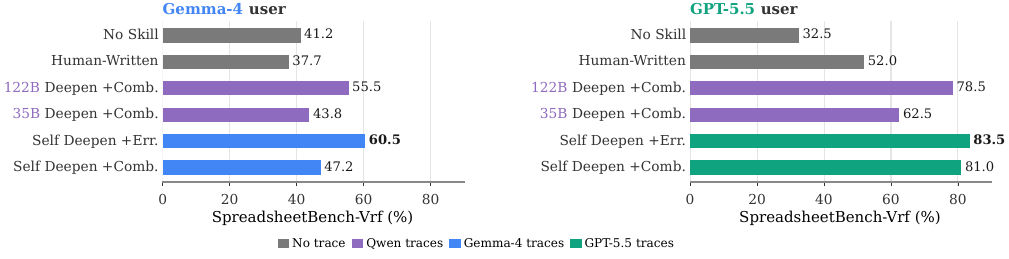}
    \caption{Cross-family generalization on SpreadsheetBench. Gemma-4-31B-it and GPT-5.5-high each (i)~evolve a skill from their own traces and (ii)~run Qwen3.5-authored skills.}
    \label{fig:model_family_generalization}
\end{figure}

We test generalization across model families with Gemma-4-31B-it and GPT-5.5-high in two settings: each model (i)~deepens the human-written \texttt{xlsx} skill from its own traces via \systemname{}, and (ii)~runs the Qwen3.5-122B/35B +Combined deepened skills. \cref{fig:model_family_generalization} shows that both models self-improve from their own traces and also benefit from Qwen3.5-authored skills, so \systemname{} generalizes across families. \cref{app:model_family_cross_trace} reports additional results and implementation details.

\begin{table}[!t]
\centering
\begin{minipage}[t]{0.48\linewidth}
\centering
\small
\setlength{\tabcolsep}{2.2pt}
\caption{Math transfer to held-out DAPO and AIME. D-Test is DAPO-Math-Test pass rate; AIME is avg@8.}
\label{tab:math}
\resizebox{\linewidth}{!}{%
\begin{tabular}{@{}l cc cc@{}}
\toprule
& \multicolumn{2}{c}{\textit{Skill User: 122B}}
& \multicolumn{2}{c}{\textit{Skill User: 35B}} \\
\cmidrule(lr){2-3}\cmidrule(lr){4-5}
\textbf{Condition}
    & \textbf{D-Test}$\uparrow$
    & \textbf{AIME}$\uparrow$
    & \textbf{D-Test}$\uparrow$
    & \textbf{AIME}$\uparrow$ \\
\midrule
\multicolumn{5}{l}{\textit{Reference (absolute scores)}} \\
\quad No Skill
    & 91.0 & 90.4 & 89.0 & 83.3 \\
\midrule
\multicolumn{5}{l}{\textit{Skill Author: Qwen3.5-122B-A10B}} \\[2pt]
\multicolumn{5}{l}{\quad\textit{Creation (init: No Skill)}} \\
\quad\quad +Error
    & \cellcolor{Green3!60}\textbf{+4.0} & \cellcolor{Green3!60}\textbf{+2.9} & \cellcolor{Green3!50}+5.0 & \cellcolor{Green3!55}+5.0 \\
\quad\quad +Success
    & \cellcolor{Green3!30}+2.0 & \cellcolor{Green3!8}+0.4 & \cellcolor{Green3!20}+2.0 & \cellcolor{Green3!50}+4.6 \\
\quad\quad +Combined
    & \cellcolor{Green3!45}+3.0 & \cellcolor{Red1!25}-1.2 & \cellcolor{Green3!60}\textbf{+6.0} & \cellcolor{Green3!60}\textbf{+5.5} \\[2pt]
\midrule
\multicolumn{5}{l}{\textit{Skill Author: Qwen3.5-35B-A3B}} \\[2pt]
\multicolumn{5}{l}{\quad\textit{Creation (init: No Skill)}} \\
\quad\quad +Error
    & \cellcolor{Green3!45}+3.0 & \cellcolor{Green3!27}+1.3 & \cellcolor{Green3!40}+4.0 & \cellcolor{Green3!5}+0.5 \\
\quad\quad +Success
    & \cellcolor{Red1!20}-1.0 & \cellcolor{Red1!25}-1.2 & +0.0 & \cellcolor{Red1!13}-1.2 \\
\quad\quad +Combined
    & \cellcolor{Green3!15}+1.0 & \cellcolor{Red1!8}-0.4 & \cellcolor{Green3!10}+1.0 & +0.0 \\
\bottomrule
\end{tabular}
}%

\end{minipage}\hfill
\begin{minipage}[t]{0.48\linewidth}
\centering
\small
\setlength{\tabcolsep}{2.2pt}
\caption{DocVQA transfer from trajectory-distilled visual-reasoning skills. ANLS is similarity; Acc is ANLS\,$\ge$\,0.5.}
\label{tab:vqa}
\resizebox{\linewidth}{!}{%
\begin{tabular}{@{}l cc cc@{}}
\toprule
& \multicolumn{2}{c}{\textit{Skill User: 122B}}
& \multicolumn{2}{c}{\textit{Skill User: 35B}} \\
\cmidrule(lr){2-3}\cmidrule(lr){4-5}
\textbf{Condition}
    & \textbf{ANLS}$\uparrow$
    & \textbf{Acc}$\uparrow$
    & \textbf{ANLS}$\uparrow$
    & \textbf{Acc}$\uparrow$ \\
\midrule
\multicolumn{5}{l}{\textit{Reference (absolute scores)}} \\
\quad No Skill
    & 0.6300 & 70.27 & 0.6582 & 73.17 \\
\midrule
\multicolumn{5}{l}{\textit{Skill Author: Qwen3.5-122B-A10B}} \\[2pt]
\multicolumn{5}{l}{\quad\textit{Creation (init: No Skill)}} \\
\quad\quad +Error
    & \cellcolor{Green3!46}+0.1949 & \cellcolor{Green3!48}+17.69 & \cellcolor{Green3!47}+0.1974 & \cellcolor{Green3!45}+16.64 \\
\quad\quad +Success
    & \cellcolor{Green3!39}+0.1639 & \cellcolor{Green3!40}+14.89 & \cellcolor{Green3!40}+0.1668 & \cellcolor{Green3!38}+14.23 \\
\quad\quad +Combined
    & \cellcolor{Green3!60}\textbf{+0.2534} & \cellcolor{Green3!60}\textbf{+22.25} & \cellcolor{Green3!49}+0.2049 & \cellcolor{Green3!46}+17.22 \\[2pt]
\midrule
\multicolumn{5}{l}{\textit{Skill Author: Qwen3.5-35B-A3B}} \\[2pt]
\multicolumn{5}{l}{\quad\textit{Creation (init: No Skill)}} \\
\quad\quad +Error
    & \cellcolor{Green3!21}+0.0884 & \cellcolor{Green3!18}+6.79 & \cellcolor{Green3!30}+0.1267 & \cellcolor{Green3!27}+10.15 \\
\quad\quad +Success
    & \cellcolor{Green3!37}+0.1572 & \cellcolor{Green3!39}+14.53 & \cellcolor{Green3!50}+0.2132 & \cellcolor{Green3!49}+18.27 \\
\quad\quad +Combined
    & \cellcolor{Green3!23}+0.0958 & \cellcolor{Green3!21}+7.73 & \cellcolor{Green3!51}\textbf{+0.2158} & \cellcolor{Green3!51}\textbf{+18.83} \\
\bottomrule
\end{tabular}
}%

\end{minipage}
\end{table}

\subsection{Math Reasoning}
\label{sec:experiments:math}
We apply \systemname{} to math domain to test its domain-agnosticism.
As in the spreadsheet setting (\cref{sec:experiments:setup}), the math agent runs in a ReAct loop with a command-line Python interpreter to write and execute code for each question.
We create skills from scratch on 400 DAPO questions \citep{yu2025dapo} and evaluate on 100 disjoint held-out DAPO questions (ID) and AIME~2026 (OOD competition mathematics; avg@8 over 30 problems), following the cross-model protocol of \cref{sec:experiments:main}.
\cref{tab:math} reports deltas from No Skill.
\textbf{\systemname{} works for math domain:} the distilled skills improve both held-out DAPO and OOD AIME rather than overfitting the source distribution.
\textbf{+Error is the most stable signal}, transferring cleanly between the 122B and 35B users.

\subsection{Visual Question Answering}
\label{sec:experiments:vqa}

To test multimodal generalization, we apply \systemname{} to DocVQA~\citep{mathew2020docvqa}, where agents answer questions over document images such as forms, tables, invoices, letters, and reports.
Again, each agent runs in a ReAct loop with a command-line + Python environment.
We use 50 DocVQA examples only for skill evolution and remove them from evaluation; the remaining 5{,}299 examples form the held-out test set.
We report the official ANLS and Accuracy (ANLS~$\geq 0.5$, \%) for both same-model use and cross-model transfer between 122B- and 35B-authored skills. Results in \cref{tab:vqa} show that \textbf{all
evolved DocVQA skills improve performance over No Skill. +Combined is the strongest setting}, giving the clearest same-model gains and positive cross-model transfer.

% Each VQA agent operates in a ReAct loop (up to 20 turns) with a code-execution tool and a benchmark-specific system prompt.
% We compare two skill conditions:
% \textbf{No Skill} (system prompt only) and
% \textbf{+Evolved} (a skill automatically produced by the closed-loop pipeline: error analysis on the evolving set $\to$ parallel MapReduce skill evolution $\to$ programmatic application).

\FloatBarrier

\section{Analysis}
\label{sec:analysis}
\FloatBarrier
\suppressfloats[t]

This section analyzes why \systemname{} works and its broader application. We first isolate its three core design choices under a shared trace pool and execution harness (\cref{sec:analysis:core_design}), then characterize the standard operating procedures it distills (\cref{sec:analysis:sops}), study how patch value composes (\cref{sec:analysis:selective_patch_aggregation}), and test transfer in broader real-world tasks (\cref{sec:analysis:broader_application}).

\subsection{Core Design Comparisons}
\label{sec:analysis:core_design}

\systemname{} rests on three design choices, which we isolate here by comparing each against its natural alternative under the same trace pool and execution harness: parallel many-to-one consolidation versus sequential editing (speed without quality loss), a single consolidated skill versus experience memory retrieval (reuse without test-time retrieval), and agentic versus single-call error analysis (more accurate root-cause patches).

\begin{table}[H]
\centering
\centering
\small
\setlength{\tabcolsep}{4pt}
\caption{Parallel consolidation versus sequential editing on SpreadsheetBench (same trace pool, +Error). Parallel outperforms Seq in effectiveness and efficiency. \cref{tab:math_seq_parallel} and \cref{tab:vqa_seq_parallel} show similar trend in math/VQA.}
\label{tab:seq_parallel}
\begin{tabular}{@{}l ccc ccc c@{}}
\toprule
& \multicolumn{3}{c}{\textit{Skill User: 122B}}
& \multicolumn{3}{c}{\textit{Skill User: 35B}} & \\
\cmidrule(lr){2-4}\cmidrule(lr){5-7}
\textbf{Condition}
    & \textbf{Vrf}$\uparrow$ & \textbf{Soft}$\uparrow$ & \textbf{Hard}$\uparrow$
    & \textbf{Vrf}$\uparrow$ & \textbf{Soft}$\uparrow$ & \textbf{Hard}$\uparrow$
    & \textbf{Time}$\downarrow$ \\
\midrule
Seq-$B{=}4$           & 59.00 & 40.63 & 20.63 & 26.17 & 22.37 &  7.47 & $\sim$15 min \\
Seq-$B{=}1$           & 61.83 & 44.40 & 25.40 & 26.00 & \textbf{23.83} & \textbf{10.57} & $\sim$60 min \\
Parallel (ours)        & \textbf{65.83} & \textbf{46.60} & \textbf{27.43} & \textbf{27.00} & 22.20 &  8.20 & \textbf{$\sim$3 min} \\
\bottomrule
\end{tabular}

\end{table}

\myparagraph{Parallel Consolidation.}
Online skill evolution typically edits the skill as trajectory batches arrive, so later analyses depend on earlier edits.
We isolate the effect of our parallel many-to-one consolidation by comparing against two sequential baselines: Seq-$B{=}1$, which updates after every trajectory, and Seq-$B{=}4$, which updates after every four trajectories.
All conditions use error analysts only and initialize from the Human-Written skill.
\Cref{tab:seq_parallel} reports the resulting quality and wall-clock tradeoff.

Parallel consolidation is better on all 122B SpreadsheetBench metrics and on 35B Vrf; the exception is that Seq-$B{=}1$ is modestly higher on 35B Soft and Hard.
This small quality tradeoff comes with a large efficiency gap: parallel produces the skill in about 3~min, compared with 15~min for Seq-$B{=}4$ and 60~min for Seq-$B{=}1$.
Math reasoning and DocVQA show the same broad trend that parallel consolidation preserves or improves quality while being much faster; \cref{app:parallel_consolidation_details} gives the latency analysis and similar comparisons on math/VQA.

\Needspace{0.40\textheight}
\myparagraph{Holistic Skill vs.\ Retrieval.}
Past-experience retrieval is a common paradigm for reusing agent trajectories: experience-memory systems store reflections, memories, or procedural lessons from previous executions and retrieve relevant items at inference time \citep{shinn2023reflexionlanguageagentsverbal,wang2023voyageropenendedembodiedagent,ouyang2026reasoningbankscalingagentselfevolving,fang2026mempexploringagentprocedural,wang2024agentworkflowmemory,liu2025contextualexperiencereplayselfimprovement}.

\begin{wraptable}{r}{0.50\linewidth}
\centering
\centering
\small
\setlength{\tabcolsep}{4pt}
\caption{\systemname{} outperforms ReasoningBank-style retrieval on SpreadsheetBench (same trajectory pool). \cref{tab:math_retrieval} and \cref{tab:vqa_retrieval} show similar trend in math/VQA.}
\label{tab:reasoning_bank}
\begin{tabular}{@{}l l ccc@{}}
\toprule
\textbf{Setting}
    & \textbf{User}
    & \textbf{Vrf}$\uparrow$
    & \textbf{Soft}$\uparrow$
    & \textbf{Hard}$\uparrow$ \\
\midrule
\multirow{2}{*}{\makecell[l]{ReasoningBank\\\citep{ouyang2026reasoningbankscalingagentselfevolving}}}
    & 122B & 56.00 & 40.10 & 21.30 \\
    & 35B & 20.50 & 17.30 & 4.97 \\
\midrule
\multirow{2}{*}{\makecell[l]{Human-Written\\+Combined (ours)}}
    & 122B & \textbf{69.83} & \textbf{47.17} & \textbf{29.53} \\
    & 35B & \textbf{29.67} & \textbf{18.80} & \textbf{5.73} \\
\bottomrule
\end{tabular}

\end{wraptable}
We instantiate this paradigm with ReasoningBank \citep{ouyang2026reasoningbankscalingagentselfevolving}: following its original setting, we store lessons from both success and failure trajectories and retrieve the top-1 memory with Qwen3-Embedding-8B.
We compare it against +Combined, which uses the same trajectory pool. Results on same-model Deepening are shown in \cref{tab:reasoning_bank}. +Combined is consistently better than ReasoningBank.
This supports the advantage of consolidating trajectory evidence into a compact skill rather than retrieving isolated memories at test time. We exclude OOD evaluations, as retrieval is not applicable for WikiTQ and HiTab whose queries are too semantically distant from the queries of SpreadsheetBench.
% We attribute the gain to two factors: the skill integrates recurring lessons into the agent's standing instructions, while retrieval remains sensitive to surface similarity between the current query and stored memory keys.

\WFclear

\begin{figure}[t]
\centering
\begin{minipage}[t]{0.48\linewidth}
\centering
\includegraphics[width=0.86\linewidth,height=0.68\linewidth]{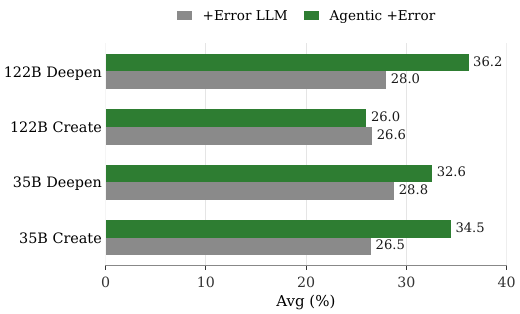}
\caption{Agentic +Error versus single-call +Error~LLM on Avg, the balance of ID and OOD performance (same as Avg in \cref{tab:main_v1}). The agentic analyst inspects artifacts and validates fixes, whereas +Error~LLM reads only the execution log.}
\label{fig:agentic_ablation_avg}
\end{minipage}\hfill
\begin{minipage}[t]{0.48\linewidth}
\centering
\begingroup
\setlength{\columnwidth}{0.82\linewidth}%
\begin{tikzpicture}
\begin{axis}[
    width=\columnwidth,
    height=0.68\columnwidth,
    xmin=0, xmax=10,
    ymin=34, ymax=72,
    xlabel={Iteration},
    ylabel={SpreadsheetBench-Vrf (\%)},
    xtick={0,1,2,3,4,5,6,7,8,9,10},
    ytick={35,45,55,65},
    tick label style={font=\scriptsize},
    label style={font=\small},
    grid=major,
    grid style={black!8},
    axis line style={black!45},
    legend style={
        at={(0.5,-0.31)},
        anchor=north,
        draw=none,
        fill=none,
        font=\scriptsize,
        legend columns=2,
        /tikz/every even column/.append style={column sep=3pt}
    },
]
\addplot+[mark=o, thick, color=blue!70!black] coordinates {
    (0,48.33) (1,41.33) (2,40.83) (3,36.67) (4,44.33)
    (5,44.50) (6,49.50) (7,47.00) (8,52.83) (9,53.33) (10,55.17)
};
\addlegendentry{Greedy +Success}

\addplot+[mark=square*, thick, color=orange!85!black] coordinates {
    (0,48.33) (1,62.67) (2,64.17) (3,65.67) (4,65.17)
    (5,65.67) (6,60.83) (7,62.50) (8,62.67) (9,64.17) (10,63.33)
};
\addlegendentry{Greedy +Error}

\addplot+[mark=triangle*, thick, color=green!55!black] coordinates {
    (0,48.33) (1,56.83) (2,66.17) (3,62.50) (4,60.17)
    (5,62.17) (6,58.50) (7,59.00) (8,59.50) (9,59.17) (10,59.17)
};
\addlegendentry{Greedy +Combined}

\addplot+[no markers, dashed, very thick, color=blue!70!black] coordinates {
    (0,66.33) (10,66.33)
};
\addlegendentry{\systemname{} +Success}

\addplot+[no markers, dashed, very thick, color=orange!85!black] coordinates {
    (0,65.83) (10,65.83)
};
\addlegendentry{\systemname{} +Error}

\addplot+[no markers, dashed, very thick, color=green!55!black] coordinates {
    (0,69.83) (10,69.83)
};
\addlegendentry{\systemname{} +Combined}
\end{axis}
\end{tikzpicture}
\endgroup
\caption{Greedy selective patch aggregation on SpreadsheetBench-Vrf (pass rate, \%). Each curve adds one validation-selected patch per iteration ($x$-axis) from the Error, Success, or Combined pool; the flat line is full \systemname{} aggregation of all patches.}
\label{fig:greedy_patch_selection_vrf}
\end{minipage}
\end{figure}
\myparagraph{Agentic Error Analysis.}
Related work derives transferable lessons or skills from error trajectories via a single non-interactive LLM call \citep{ouyang2026reasoningbankscalingagentselfevolving,xia2026skillrlevolvingagentsrecursive,yang2026autoskillexperiencedrivenlifelonglearning,jiang2026xskillcontinuallearningexperience}. We ablate this design with \textbf{+Error~LLM}, where one LLM call reads each failed trajectory and proposes a patch without inspecting artifacts, querying ground truth, or validating fixes; \cref{fig:agentic_ablation_avg} compares it with our agentic error analyst. Agentic +Error achieves higher Avg than +Error~LLM in most settings, showing that interactive diagnosis is more useful than log-only patch generation; \cref{app:agentic_ablation_full} reports and discusses the full per-dataset results in \cref{tab:agentic_ablation}. The qualitative analysis in \cref{app:qual_agentic_vs_llm} shows the mechanism: artifact access and fix validation help the agentic analyst identify root causes more precisely.

\myparagraph{Apples-to-apples vs.\ head-to-head comparison.}
The above three comparisons are apples-to-apples, isolating each core design without the data, model, harness, and engineering confounders that complicate whole-system evaluation. As a complementary view, \cref{app:prior_systems} reports a direct head-to-head against three concurrent systems (XSkill \citep{jiang2026xskillcontinuallearningexperience}, EvoSkill \citep{alzubi2026evoskillautomatedskilldiscovery}, and SkillGen \citep{ma2026skillgenverifiedinferencetime}) under a shared open model and benchmark, where \systemname{} still shows its advantage.

\subsection{SoPs Learned}
\label{sec:analysis:sops}

Many-to-one merging keeps the operations that recur across trajectories and turns them into standard operating procedures (SoPs), rather than collecting one-off tricks.
Across the 323 map patches from the 122B Deepening +Combined run, four learned SoPs dominate, each cited by between $16\%$ and $55\%$ of patches (a single patch can cite more than one SoP), including: recalculating and reading back formulas after writes, verifying target cells before submitting, and deleting rows in a corruption-safe order.
Task-specific quirks are routed to on-demand \texttt{references/} files, keeping the SKILL.md focused on broadly reusable procedures (Details in \cref{app:sops}).

\subsection{Selective Patch Aggregation}
\label{sec:analysis:selective_patch_aggregation}

\systemname{} aggregates all learned patches into the target skill, avoiding the cost of per-patch validation. If patches vary in quality, can selecting a subset do better? We test two selectors against full aggregation: \textbf{greedy top-1}, which at each iteration adds the locally best patch according to validation on evolving set, and \textbf{Bayesian optimization (BO)}, which searches binary patch-inclusion vectors and reuses validation scores from previously tried subsets to bias later proposals toward promising combinations. Both use the 200-task evolving set (\cref{sec:experiments:main}) and a 32-task validation set from evolving set; \cref{app:selective_patch_aggregation} gives the full algorithmic details.

\myparagraph{Greedy rises quickly but plateaus below full aggregation.}
\cref{fig:greedy_patch_selection_vrf} shows greedy +Error and +Combined rising for a few iterations and then plateauing, with +Success dipping and only partially recovering; every greedy curve stays below full \systemname{} aggregation.
Two mechanisms drive the plateau (detailed in \cref{app:qual_patch_selection}).
(1) \emph{patch-irrelevant regression}: the patches flip some target tasks from wrong to correct, but their side effects also flip previously-correct tasks from correct to wrong, so net accuracy stalls.
(2) \emph{semantic overlap}: recurring errors lead the pipeline to propose patches that repeatedly target the same behaviors (e.g., formula recalculation, validation checklists), so a new patch largely restates existing guidance and adds little marginal gain.
Patch value is therefore combinatorial rather than a sum of singletons, which makes optimizing the combined effect of patches necessary.

\myparagraph{BO is useful but computationally heavy.}
BO searches patch subsets directly, so it can model complementarity and interference that greedy misses.
\cref{tab:bo_patch_selection} shows that, compared with applying all +Error patches, BO improves most selected metrics, especially the SpreadsheetBench columns.
The cost is validation: each candidate subset must be materialized as a skill and run on the validation set before it can inform the posterior, and larger patch universes increase this cost quickly.
We therefore view BO as useful when the target distribution and validation budget are known, rather than as a replacement for the default full aggregation in \systemname{}.

\begin{table}[t]
\centering
\setlength{\tabcolsep}{5pt}
\caption{BO-selected +Error patches versus applying all of them (No Selection), in the 122B Deepening spreadsheet setting. Evolve is the accuracy on evolving-set.}
\begin{tabular}{lcccccc}
\toprule
\textbf{Skill} & \textbf{Evolve} & \textbf{Vrf} & \textbf{Soft} & \textbf{Hard} & \textbf{WikiTQ} & \textbf{HiTab} \\
\midrule
No Selection (+Error) & 68.00 & 65.83 & 46.60 & 27.43 & 76.30 & \textbf{43.45} \\
BO Selection (+Error) & \textbf{72.83} & \textbf{69.83} & \textbf{47.27} & \textbf{28.90} & \textbf{77.51} & 43.38 \\
\bottomrule
\end{tabular}

\label{tab:bo_patch_selection}
\end{table}

\subsection{Broader Application}
\label{sec:analysis:broader_application}
We test broader application of \systemname{} in PDF extraction, PPTX editing, and DOCX editing. These settings use Anthropic's official \texttt{pdf}, \texttt{pptx}, and \texttt{docx} skills as starting points \citep{anthropic2026documentskills} and retain the same execution style as the spreadsheet experiments: agents operate over files with command-line tools and are scored by task-specific local verifiers.

Across the three domains, evolution on source traces improve performance on separate held-out targets. For PDF, VRDU traces transfer to VAREX, raising pass rate from 76.9\% to 85.3\% \citep{wang2023vrdu,barzelay2026varex}. For PPTX, TSBench traces transfer to a deck-disjoint TSBench OOD split, improving 72.5\% to 88.8\% \citep{jung2026talktoyourslides}. For DOCX, evolving on synthetic tasks that cover generic DOCX operations transfers to the OfficeBench DOCX subset, improving 79.7\% to 87.5\% \citep{wang2024officebench}. \cref{app:broader_application} provides setup and implementation details for each domain.

\section{Related Work}
\label{sec:related}

\myparagraph{Agent skills.}
Agent skills package task procedures, domain knowledge, and operational guardrails into loadable artifacts \citep{anthropic2026skills}, but ecosystems and benchmarks show the abstraction is delicate: focused, well-matched skills improve performance, while broad, stale, or mismatched ones can distract or even hurt the agent \citep{zhou2026comprehensivesurveyagentskills,li2026skillsbenchbenchmarkingagentskills,han2026sweskillsbenchagentskillsactually,li2026singleagentskillsreplacemultiagent,li2026organizingorchestratingbenchmarkingagent,liang2026skillnetcreateevaluateconnect}. \systemname{} keeps this view of skills as portable SoPs, but asks a narrower question: how to compress broad execution evidence into guidance that stays useful across models and task distributions.

\myparagraph{Experience memory for agent self-evolution.}
Another line improves agents by storing execution experience for reuse, from verbal reflection and accumulated behaviors to retrieval, procedural, workflow, or replay memories queried at test time \citep{shinn2023reflexionlanguageagentsverbal,wang2023voyageropenendedembodiedagent,ouyang2026reasoningbankscalingagentselfevolving,fang2026mempexploringagentprocedural,wang2024agentworkflowmemory,qian2024investigateconsolidateexploitgeneralstrategyintertask,nottingham2024skillsetoptimizationreinforcing,liu2025contextualexperiencereplayselfimprovement}. These methods share our premise that experience contains reusable structure but retain an external memory or retrieval module, whereas \systemname{} distills many local observations into one static skill directory. It favors inductive compression over nearest-neighbor reuse and avoiding test-time dependence on retrieval quality.

\myparagraph{Skill and policy evolution.}
The closest concurrent line evolves skills from agent interaction or guided refinement
\citep{zheng2025skillweaverwebagentsselfimprove,yang2026autoskillexperiencedrivenlifelonglearning,jiang2026xskillcontinuallearningexperience,alzubi2026evoskillautomatedskilldiscovery,zhou2026mementoskillsletagentsdesign,ma2026skillgenverifiedinferencetime,anthropic2026skillcreatorconversation},
while a broader family co-evolves policies, skills, or agent loops online through skill-augmented reinforcement learning, intrinsic skill evolution, or meta-learning
\citep{xia2026skillrlevolvingagentsrecursive,li2026ariseagentreasoningintrinsic,xia2026metaclawjusttalk}.
\systemname{} asks a complementary question: whether traces collected with one model can be compressed into a static skill directory that directly benefits other models and tasks. Instead of maintaining a test-time memory, retrieval index, or updated policy, it consolidates trajectory-local evidence into reusable SoPs that can be loaded unchanged. Thus, our focus is not adaptive reuse within a particular agent loop, but portability of the learned artifact across deployment settings.
% conclusion_v2.tex — Trajectory-grounded distillation framing; generalization as central finding.

\section{Conclusion}
\label{sec:conclusion}

We introduced \systemname, a framework that distills agent execution traces into a portable skill: parallel analyst sub-agents propose targeted patches from disjoint trajectory batches, and a consolidation step merges them at once into one declarative skill directory. Skills distilled from a single model's traces transfer across model scales, families, and OOD tasks. \systemname{} also exhibits strong applicability in various domains.

\section*{Limitations}
\systemname{} applies all consolidated patches by default; selecting a higher-quality subset with Bayesian optimization can help (\cref{sec:analysis:selective_patch_aggregation}), but it is costly. A reliable selection signal needs a large, representative validation set and BO must materialize and score a new skill for every candidate subset, so cost grows quickly with both the validation set and the patch universe. Ideally, using the whole evolving set for validation would better reflect patch quality than our sampled validation with only 32 questions, but it would be much more computationally expensive. We therefore leave a more thorough exploration of combinatorial patch selection to future work.

\section*{Ethics Statement}

We use publicly available datasets, which have no data privacy issues. All artifacts we use are under licenses allowing research usage. Human annotation in qualitative analyses were conducted by the authors of this paper. We do not identify any other ethical risks associated with this study. AI coding tools (Codex/Claude Code) are used to assist with coding. We confirm that all such coding is under careful human supervision. Unit tests are implemented to avoid hacking or unaligned behavior. We will also open-source the code for full reproducibility. We also leverage ChatGPT for grammar check and fix, fully supervised by human authors.

\bibliography{main_v2}
\bibliographystyle{conference}

\appendix
% \clearpage
\section{Experimental Details}
\label{app:experimental_details}

\myparagraph{Random seeds and compute.}
Unless otherwise noted, all reported results are averaged over three random seeds: 41, 42, and 43.
Experiments are run on nodes with 8 NVIDIA A100 GPUs. We spent roughly a total of 20,000 GPU hours for all experiments and exploration.

\begin{table*}[t]
\centering
\setlength{\tabcolsep}{3.5pt}
\caption{Standard deviations across seeds 41, 42, and 43 for the main spreadsheet results in \cref{tab:main_v1}. Reference rows are standard deviations of absolute scores; evolved rows are standard deviations of the paired deltas defined in \cref{tab:main_v1}. The summary metric Avg is stable (standard deviation $\le 3.4$ in every row), and the mean gains in \cref{tab:main_v1} stay large relative to the per-cell spread, so the comparisons are robust. The higher variance in individual columns (Vrf and the OOD WikiTQ/HiTab transfer) is expected rather than a sign of instability: each score aggregates long agentic rollouts that often exceed 30 turns of environment interaction, so single-benchmark run-to-run stochasticity accumulates while averaging out in Avg. For conciseness, we omit additional standard-deviation tables for the paper's other results, whose deviations are similarly small: std ranges for spreadsheet tasks are similar to this table. Std ranges of DAPO-Test, AIME'26, and DocVQA ANLS are 1.1-2.4, 1.8-3.4, and 0.034-0.139 correspondingly.}
\label{tab:main_std}
\resizebox{\textwidth}{!}{%
\begin{tabular}{@{}l ccccc ccccc c@{}}
\toprule
& \multicolumn{5}{c}{\textit{Skill User: Qwen3.5-122B-A10B}}
& \multicolumn{5}{c}{\textit{Skill User: Qwen3.5-35B-A3B}}
& \\
\cmidrule(lr){2-6}\cmidrule(lr){7-11}
& \multicolumn{3}{c}{\textit{SpreadsheetBench}} & \multicolumn{2}{c}{\textit{OOD}}
& \multicolumn{3}{c}{\textit{SpreadsheetBench}} & \multicolumn{2}{c}{\textit{OOD}}
& \\
\cmidrule(lr){2-4}\cmidrule(lr){5-6}\cmidrule(lr){7-9}\cmidrule(lr){10-11}
\textbf{Condition}
    & \textbf{Vrf}$\uparrow$ & \textbf{Soft}$\uparrow$ & \textbf{Hard}$\uparrow$ & \textbf{WikiTQ}$\uparrow$ & \textbf{HiTab}$\uparrow$
    & \textbf{Vrf}$\uparrow$ & \textbf{Soft}$\uparrow$ & \textbf{Hard}$\uparrow$ & \textbf{WikiTQ}$\uparrow$ & \textbf{HiTab}$\uparrow$
    & \textbf{Avg}$\uparrow$ \\
\midrule
\multicolumn{12}{l}{\textit{Baseline (absolute scores)}} \\
\quad No Skill
    & 3.75 & 0.90 & 0.70 & 0.81 & 0.67 & 1.80 & 2.77 & 1.25 & 1.27 & 5.28 & 0.42 \\
\quad Human-Written
    & 8.10 & 1.39 & 0.40 & 0.65 & 0.71 & 2.93 & 0.57 & 0.40 & 0.36 & 2.68 & 0.46 \\
\quad Parametric
    & 0.58 & 1.04 & 0.75 & 6.52 & 8.93 & 1.44 & 1.56 & 0.50 & 13.32 & 18.31 & 3.28 \\
\midrule
\multicolumn{12}{l}{\textit{Skill Author: Qwen3.5-122B-A10B}} \\[2pt]
\multicolumn{12}{l}{\quad\textit{Deepening (Delta from init: Human-Written)}} \\
\quad\quad +Error
    & 4.82 & 1.10 & 0.78 & 0.78 & 0.60 & 6.76 & 3.36 & 2.06 & 2.62 & 4.16 & 1.02 \\
\quad\quad +Success
    & 12.53 & 0.75 & 1.35 & 1.68 & 0.75 & 3.69 & 3.78 & 2.23 & 1.43 & 4.01 & 1.43 \\
\quad\quad +Combined
    & 9.53 & 1.68 & 0.61 & 1.22 & 0.85 & 5.92 & 2.92 & 1.73 & 2.53 & 2.74 & 1.57 \\[2pt]
\multicolumn{12}{l}{\quad\textit{Creation (Delta from init: Parametric)}} \\
\quad\quad +Error
    & 1.53 & 1.10 & 0.55 & 7.07 & 6.98 & 3.62 & 1.63 & 1.22 & 13.30 & 18.93 & 3.24 \\
\quad\quad +Success
    & 1.89 & 0.91 & 1.45 & 5.25 & 9.60 & 4.86 & 2.39 & 1.76 & 14.66 & 19.38 & 3.00 \\
\quad\quad +Combined
    & 3.04 & 0.90 & 1.10 & 6.07 & 9.69 & 3.04 & 1.48 & 1.13 & 13.52 & 19.72 & 2.99 \\[2pt]
\midrule
\multicolumn{12}{l}{\textit{Skill Author: Qwen3.5-35B-A3B}} \\[2pt]
\multicolumn{12}{l}{\quad\textit{Deepening (init: Human-Written)}} \\
\quad\quad +Error
    & 11.50 & 1.31 & 2.31 & 1.40 & 0.63 & 5.11 & 3.00 & 1.12 & 0.67 & 1.46 & 1.10 \\
\quad\quad +Success
    & 15.81 & 4.63 & 2.42 & 1.69 & 0.84 & 0.76 & 2.47 & 1.16 & 1.18 & 2.90 & 1.32 \\
\quad\quad +Combined
    & 9.00 & 2.14 & 1.33 & 1.94 & 0.85 & 0.50 & 3.62 & 2.54 & 3.27 & 2.71 & 0.88 \\[2pt]
\multicolumn{12}{l}{\quad\textit{Creation (init: Parametric)}} \\
\quad\quad +Error
    & 2.18 & 1.04 & 0.58 & 6.96 & 8.84 & 1.76 & 1.77 & 1.08 & 14.60 & 17.83 & 3.36 \\
\quad\quad +Success
    & 1.61 & 1.00 & 0.90 & 6.94 & 8.64 & 4.51 & 1.10 & 1.19 & 14.24 & 19.02 & 3.21 \\
\quad\quad +Combined
    & 2.75 & 1.02 & 1.28 & 6.37 & 8.19 & 5.51 & 1.62 & 1.91 & 10.98 & 19.64 & 3.16 \\[2pt]
\bottomrule
\end{tabular}
}%
\end{table*}

\myparagraph{Spreadsheet data and scoring.}
SpreadsheetBench Verified is split into 200 evolution problems and 200 held-out test problems; no held-out sample is used during skill evolution.
We additionally report Soft, the sub-problem pass rate, and Hard, where all sub-problems must pass, on full SpreadsheetBench, from which we remove every evolving-set problem so that no problem seen during skill evolution is scored at test time.
For OOD transfer, WikiTableQuestions is converted into spreadsheet tasks over compositional Wikipedia tables, and HiTab is converted into spreadsheet tasks that require hierarchical indexing plus implicit calculation and semantic reasoning.
The evaluation sets contain 200 SpreadsheetBench-Verified held-out examples, 2{,}529 full SpreadsheetBench examples for Soft/Hard (all evolving-set problems removed), 2{,}810 WikiTableQuestions examples, and 1{,}585 HiTab examples.

\myparagraph{Baseline skills.}
Human-Written is Anthropic's official \texttt{xlsx} skill, and Parametric is an \texttt{xlsx-basic} seed generated by prompting Qwen3.5-122B-A10B from parametric knowledge alone, with no trajectory grounding.

\myparagraph{Model serving and evolution protocol.}
The two main author/user models are Qwen3.5-122B-A10B and Qwen3.5-35B-A3B.
We use the Hugging Face checkpoints \href{https://huggingface.co/Qwen/Qwen3.5-122B-A10B}{Qwen/Qwen3.5-122B-A10B} and \href{https://huggingface.co/Qwen/Qwen3.5-35B-A3B}{Qwen/Qwen3.5-35B-A3B}.
Both are instruct/think hybrid MoE models: we use instruct mode for multi-turn ReAct-style agents and thinking mode for single-call steps such as hierarchical merging, success analysis, and patch conversion.
Models are served with vLLM \citep{kwon2023efficientmemorymanagementlarge}; for reasoning-mode calls, we follow Qwen's official recommended decoding settings.
Stage~1 generates one trajectory per problem; Stage~2 runs 128 sub-agents in parallel with merge batch size 32.
We do not impose a separate tool-call budget.

\section{Additional Analysis Results}
\label{app:parallel_consolidation_details}

\subsection{Latency Analysis}
\label{app:parallel_latency}

With $W{=}128$ workers and $N{\approx}70$ error lessons, all analysts execute in a single parallel round.
With merge batch size $B_\text{merge}{=}32$, the hierarchical merge adds only $\lceil\log_{B_\text{merge}} N\rceil{\approx}2$ further sequential rounds, one per merge layer, yielding ${\approx}3$ sequential LLM-call rounds in total.
The sequential baselines require $N$ and $\lceil N/B\rceil$ rounds respectively, since each skill edit depends on the preceding one.
In practice this translates to 3~min for parallel consolidation, 60~min for Seq-$B{=}1$ ($20{\times}$ slower), and 15~min for Seq-$B{=}4$ ($5{\times}$ slower), with the gap scaling linearly in $N$.
All times are wall-clock skill-generation times measured on the same 8-GPU A100 node configuration.

\subsection{Math Reasoning}
\label{app:math_ablations}

\Cref{tab:math_seq_parallel,tab:math_retrieval} extend the main analysis ablations to math reasoning.
We use the +Error setting for these math ablations because it is the strongest math setting in \cref{tab:math}.
Parallel consolidation is best or tied for best on all reported math metrics while also having the lowest skill-generation wall time.

\begin{table}[t]
\begin{minipage}[t]{0.49\linewidth}
\centering
\small
\setlength{\tabcolsep}{4pt}
\caption{Parallel consolidation versus sequential editing on math reasoning (same trace budget). D-Test/AIME are as in \cref{tab:math} (122B and 35B users).}
\label{tab:math_seq_parallel}
\resizebox{\linewidth}{!}{%
\begin{tabular}{@{}l cccc c@{}}
\toprule
\textbf{Condition}
    & \multicolumn{2}{c}{\textit{Runner: 122B}}
    & \multicolumn{2}{c}{\textit{Runner: 35B}}
    & \\
\cmidrule(lr){2-3}\cmidrule(lr){4-5}
    & \textbf{D-Test}$\uparrow$
    & \textbf{AIME}$\uparrow$
    & \textbf{D-Test}$\uparrow$
    & \textbf{AIME}$\uparrow$
    & \textbf{Time}$\downarrow$ \\
\midrule
No Skill      & 92.0 & 90.4 & 89.0 & 83.3 & 0 \\
Seq-$B{=}4$   & 93.0 & \textbf{91.7} & 89.0 & 68.3 & 3.8 min \\
Seq-$B{=}1$   & 94.0 & 85.8 & 90.0 & 81.7 & 25.9 min \\
Parallel      & \textbf{95.0} & \textbf{91.7} & \textbf{94.0} & \textbf{88.3} & \textbf{2.0 min} \\
\bottomrule
\end{tabular}
}%

\end{minipage}\hfill
\begin{minipage}[t]{0.49\linewidth}
\centering
\small
\setlength{\tabcolsep}{4pt}
\caption{\systemname{} outperforms ReasoningBank retrieval on math reasoning. D-Test/AIME are as in \cref{tab:math}.}
\label{tab:math_retrieval}
\resizebox{\linewidth}{!}{%
\begin{tabular}{@{}l cccc@{}}
\toprule
\textbf{Condition}
    & \multicolumn{2}{c}{\textit{Runner: 122B}}
    & \multicolumn{2}{c}{\textit{Runner: 35B}} \\
\cmidrule(lr){2-3}\cmidrule(lr){4-5}
    & \textbf{D-Test}$\uparrow$
    & \textbf{AIME}$\uparrow$
    & \textbf{D-Test}$\uparrow$
    & \textbf{AIME}$\uparrow$ \\
\midrule
No Skill & 92.0 & 90.4 & 89.0 & 83.3 \\
ReasoningBank (error-only) & 94.0 & 90.8 & 91.0 & 80.8 \\
ReasoningBank (combined) & 94.0 & 90.4 & 91.0 & 80.4 \\
\systemname{} +Error (ours) & \textbf{95.0} & \textbf{91.7} & \textbf{94.0} & \textbf{88.3} \\
\bottomrule
\end{tabular}
}%

\end{minipage}
\end{table}

\subsection{DocVQA}
\label{app:vqa_ablations}

\Cref{tab:vqa_seq_parallel,tab:vqa_retrieval} extend the main DocVQA results to sequential and retrieval-memory comparisons.
We use the +Combined setting for these DocVQA ablations because it is the strongest DocVQA setting in \cref{tab:vqa}.
\Cref{tab:vqa_seq_parallel} isolates the 122B combined-protocol batch ablation, where all non-baseline rows use the same 25 failure and 25 success trajectories and differ only in sequential batch size versus parallel consolidation. These 50 DocVQA examples are used only for evolution and are excluded from evaluation; all DocVQA scores are computed on the remaining 5{,}299 held-out examples. The efficiency gap between parallel and sequential paradigms will grow with the number of traces as analyzed in \cref{app:parallel_latency}.

\begin{table}[t]
\begin{minipage}[t]{0.49\linewidth}
\centering
\small
\setlength{\tabcolsep}{4pt}
\caption{Parallel consolidation versus sequential editing on DocVQA using the +Combined trace pool. ANLS/Acc are as in \cref{tab:vqa}.}
\label{tab:vqa_seq_parallel}
\begin{tabular}{@{}l cc c@{}}
\toprule
\textbf{Setting}
    & \multicolumn{2}{c}{\textit{Runner: 122B}}
    & \\
\cmidrule(lr){2-3}
    & \textbf{ANLS}$\uparrow$
    & \textbf{Acc}$\uparrow$
    & \textbf{Time}$\downarrow$ \\
\midrule
No Skill      & 0.6300 & 70.27 & 0 \\
Seq-$B{=}4$   & 0.8674 & 90.80 & 5.2\,min \\
Seq-$B{=}1$   & 0.8694 & 91.05 & 35.9\,min \\
Parallel      & \textbf{0.8833} & \textbf{92.52} & \textbf{4.8\,min} \\
\bottomrule
\end{tabular}

\end{minipage}\hfill
\begin{minipage}[t]{0.49\linewidth}
\centering
\small
\setlength{\tabcolsep}{4pt}
\caption{\systemname{} outperforms ReasoningBank retrieval on DocVQA. ANLS/Acc are as in \cref{tab:vqa}.}
\label{tab:vqa_retrieval}
\resizebox{\linewidth}{!}{%
\begin{tabular}{@{}l cccc@{}}
\toprule
\textbf{Setting}
    & \multicolumn{2}{c}{\textit{Runner: 122B}}
    & \multicolumn{2}{c}{\textit{Runner: 35B}} \\
\cmidrule(lr){2-3}\cmidrule(lr){4-5}
    & \textbf{ANLS}$\uparrow$
    & \textbf{Acc}$\uparrow$
    & \textbf{ANLS}$\uparrow$
    & \textbf{Acc}$\uparrow$ \\
\midrule
No Skill                    & 0.6300 & 70.27 & 0.6582 & 73.17 \\
ReasoningBank (combined)      & 0.8668 & 90.90 & 0.8568 & 89.62 \\
\systemname{} +Combined (ours) & \textbf{0.8833} & \textbf{92.52} & \textbf{0.8740} & \textbf{92.00} \\
\bottomrule
\end{tabular}
}%

\end{minipage}
\end{table}

\Needspace{0.50\textheight}
\subsection{Cross-Model Trace Induction}
\label{app:model_family_cross_trace}

\begin{wrapfigure}[15]{r}{0.50\linewidth}
    \centering
    \includegraphics[width=\linewidth]{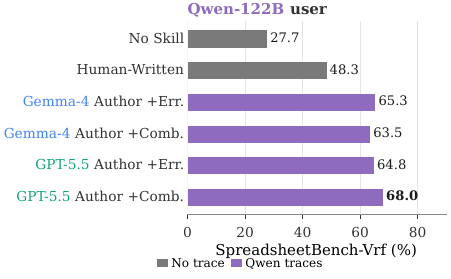}
    \caption{Cross-model trace induction on SpreadsheetBench-Vrf (pass rate, \%). Gemma-4 and GPT-5.5-high author skills from Qwen traces, which are then run by the Qwen3.5-122B user.}
    \label{fig:model_family_cross_trace}
\end{wrapfigure}

\Cref{fig:model_family_cross_trace} visualizes this cross-model trace-induction setting.
Using Qwen traces, Gemma-4-31B-it and GPT-5.5-high-authored Deepening skills improve the Qwen3.5-122B user over both No Skill (27.67\% Vrf) and Human-Written (48.33\% Vrf).
The strongest setting is GPT-5.5-high-authored Deepening +Combined at 68.00\% Vrf, showing that cross-model trace evidence can still induce meaningful portable skills.

\myparagraph{Implementation details.}
We serve Gemma-4-31B-it with vLLM \citep{kwon2023efficientmemorymanagementlarge} using the official recommended generation configuration from its Hugging Face model card (\href{https://huggingface.co/google/gemma-4-31B-it}{google/gemma-4-31B-it}); reasoning is disabled for the ReAct agent runs.
GPT-5.5-high is accessed through the official OpenAI API with only the reasoning effort set to high and all other settings left at their defaults.

\section{Qualitative Analyses}
\label{app:qualitative_analyses}

\subsection{Agentic vs.\ LLM Error Analysis}
\label{app:qual_agentic_vs_llm}

We qualitatively audit 33 shared error cases analyzed by both the agentic error analyst $\mathcal{A}^-$ and the single-call +Error~LLM baseline.
The two pipelines reach strong agreement on only 4 cases (12.1\%), while 18 cases (54.5\%) show clear disagreement about the root cause or the appropriate skill patch.
The main difference is access to evidence: $\mathcal{A}^-$ can inspect input/output artifacts, compare the submitted answer with ground truth, and validate candidate fixes, whereas +Error~LLM must infer the failure from the execution log alone.

This limitation makes the LLM-only analyzer prone to log-level false positives.
Among cases where parse-error messages appear, +Error~LLM attributes the parse error as the primary root cause in 57\% of cases, compared with 14\% for the agentic analyst.
In one representative trajectory, +Error~LLM hallucinated three distinct failure causes even though artifact evaluation showed the output was already correct.
By contrast, the agentic loop can reject such explanations after checking the generated file and rerunning the relevant validation steps.

These qualitative differences explain why the agentic patches transfer more reliably in \cref{sec:analysis:core_design}.
Rather than encoding surface symptoms from logs, $\mathcal{A}^-$ anchors patches to verified failure mechanisms: wrong target ranges, stale formula values, corrupted workbook structure, type conversion, or missing read-back verification.
The resulting edits are more likely to become domain-general guardrails and less likely to degrade ID, cross-model, or OOD settings when applied as a portable skill.

\subsection{Patch Selection Qualitative Analysis}
\label{app:qual_patch_selection}

\myparagraph{Patch-irrelevant regression.}
Later iterative patches do change agent behavior on some tasks, flipping them from wrong to correct as the patch intends.
However, the same edits also change behavior that is irrelevant to the targeted failure, flipping other, previously-correct tasks from correct to wrong.
The flips are largely off-source: across the ten-iteration greedy +Combined path (seed 41), the selected patches' source tasks have zero overlap with the evaluation split, so the previously-correct tasks a step breaks are ones the selected patch never targeted: side effects rather than targeted regressions.
Accordingly, every materialized step records substantial flips in both directions (for example, $57$ fail-to-pass against $21$ pass-to-fail at one step, and $26$ against $46$ at another), so each addition simultaneously fixes and breaks tasks. We closely inspect those flips, finding that the majority of fail-to-pass flips are patch-relevant while pass-to-fail flips are usually irrelevant
Because each greedy step both fixes and breaks tasks, the net accuracy plateaus instead of rising.
A locally best single patch therefore does not compose into a global gain: patch value depends on how additions interact, so optimizing the combinatorial effect of patches rather than their isolated marginal contributions is necessary.

\myparagraph{Semantic overlap.}
The same failure modes recur across iterations, so the pipeline repeatedly proposes semantically overlapping patches that target the same spreadsheet behaviors---formula recalculation, workbook-structure preservation, reference consolidation, and validation checklists.
The greedy path makes this concrete: across its ten iterations the selector keeps returning to a few themes---recalculation and verification (iterations 1 and 6), structure/header/sheet handling (iterations 2 and 9), and formatting/type/date handling (iterations 4, 5, 7, and 8).
A newly selected patch from an already-covered theme then largely restates guidance the skill already encodes and brings little marginal gain.

\subsection{Learned SoP Details}
\label{app:sops}
\label{app:sops_secondary}

We inspect the 323 map patches produced by the 122B Deepening +Combined run.
The four most prevalent SoPs are cited by $55.1\%$, $54.8\%$, $42.7\%$, and $16.4\%$ of the 323 patches, respectively\footnote{\systemname{} merging process tracks the original patches where the merged patch (SoPs) are induced from.}; these shares sum above $100\%$ because a single patch can cite multiple themes.

\myparagraph{Formula recalculation and write-back verification (55.1\% of patches).}
Run \texttt{recalc.py} after every formula write and reopen with \texttt{data\_only=True} to confirm evaluation; skipping this step leaves cells stale and is the single most common error mode in the run.

\myparagraph{Tool selection: \texttt{openpyxl} over \texttt{pandas.to\_excel()} (54.8\% of patches).}
Use \texttt{pandas} for read/transform logic and \texttt{openpyxl} for write-back; copy the input file to the output path first to preserve all structural anchors. \texttt{pandas.to\_excel()} silently destroys formula relationships and named ranges.

\myparagraph{Explicit read-back verification (42.7\% of patches).}
After writing, reopen the output file and confirm every target cell holds the expected value before submitting; error trajectories that fail characteristically omit this check.

\myparagraph{Structural-edit safety (16.4\% of patches).}
Delete rows in descending order to prevent index-shift corruption; copy the input workbook before editing to preserve formatting and formulas. Error trajectories document both failure modes; success trajectories confirm the protective workflow.

\myparagraph{Niche quirks are routed to \texttt{references/}.}
Low-support observations are not discarded but routed into 13 supplementary reference files rather than the main \texttt{SKILL.md}. For example, cell color extraction and FIFO vs.\ LIFO mismatch under special business logic are placed in on-demand references.
This mirrors established skill-design practice: procedural guidance flows from general to case-specific, with the main document encoding universal workflow rules and \texttt{references/} serving as an on-demand look-up layer for infrequent edge cases.
\systemname~recovers this hierarchy automatically from trajectory evidence rather than requiring manual curation.

\myparagraph{Additional moderate-support SoPs.}
The following SoPs appear in the same run with moderate support (about 3.1\%--4.6\% of patches) and are also encoded in the evolved skill.

\myparagraph{Target-range and answer-position validation (4.6\% of patches).}
Before writing, verify the exact target sheet name, cell range, and \texttt{answer\_position} field from the task metadata.
Misreading these fields --- writing to the wrong sheet or an off-by-one range --- causes silent failures that produce no error message but score zero.

\myparagraph{Datatype and datetime preservation (4.6\% of patches).}
Write dates and numeric values as native Python types, not strings.
Both \texttt{pandas} date parsing and \texttt{openpyxl} cell assignment can silently stringify datetime values; inspect each column's dtype before writing and use \texttt{openpyxl}'s native datetime assignment.

\myparagraph{Workbook structure exploration before editing (success-dominant, $\sim$4.0\% of patches).}
List all sheets, inspect row/column layout, and verify header positions before any write.
This pre-edit exploration prevents wrong-sheet and wrong-range failures and accounts for a substantial share of the 151 success-leaning patches in the run.

\section{Full Agentic Error Analysis Results}
\label{app:agentic_ablation_full}

\cref{tab:agentic_ablation} expands the main-text Avg comparison into per-dataset results for both skill authors, both evolution modes, and both skill-user models.
The main pattern is that agentic error analysis is more consistent: +Error (ours) has the higher Avg in three of the four author--mode blocks and wins most SpreadsheetBench columns.
The exception is 122B-authored Creation, where +Error~LLM is slightly higher on Avg because it does better on the OOD table benchmarks; even there, the agentic analyst remains stronger on the in-distribution SpreadsheetBench metrics.
Overall, the full table supports the main claim that artifact inspection and fix validation produce error patches that transfer more reliably than log-only analysis.

\begin{table*}[t]
\centering
\small
\setlength{\tabcolsep}{3.5pt}
\caption{Agentic error analysis (+Error) versus single-call +Error~LLM, across Deepening and Creation and all 122B/35B author--user pairs. Vrf/Soft/Hard, the OOD splits (WikiTQ/HiTab), and Avg are as in \cref{tab:main_v1}. \textbf{Bold} marks the better of the two methods per column.}
\label{tab:agentic_ablation}
\resizebox{\textwidth}{!}{%
\begin{tabular}{@{}l ccccc ccccc c@{}}
\toprule
& \multicolumn{5}{c}{\textit{Skill User: Qwen3.5-122B-A10B}}
& \multicolumn{5}{c}{\textit{Skill User: Qwen3.5-35B-A3B}}
& \\
\cmidrule(lr){2-6}\cmidrule(lr){7-11}
& \multicolumn{3}{c}{\textit{SpreadsheetBench}} & \multicolumn{2}{c}{\textit{OOD}}
& \multicolumn{3}{c}{\textit{SpreadsheetBench}} & \multicolumn{2}{c}{\textit{OOD}}
& \\
\cmidrule(lr){2-4}\cmidrule(lr){5-6}\cmidrule(lr){7-9}\cmidrule(lr){10-11}
\textbf{Condition}
    & \textbf{Vrf} & \textbf{Soft} & \textbf{Hard} & \textbf{WikiTQ} & \textbf{HiTab}
    & \textbf{Vrf} & \textbf{Soft} & \textbf{Hard} & \textbf{WikiTQ} & \textbf{HiTab}
    & \textbf{Avg} \\
\midrule
%% ── 122B Author ──────────────────────────────────────────────────────────────
\multicolumn{12}{l}{\textit{Skill Author: Qwen3.5-122B-A10B}} \\[2pt]
\multicolumn{12}{l}{\quad\textit{Deepening}} \\
\quad +Error (ours)
    & {65.83}      & {\textbf{46.60}} & {\textbf{27.43}} & {\textbf{76.30}} & {\textbf{43.45}}
    & \textbf{36.67}          & \textbf{22.47}              & \textbf{6.23}               & \textbf{18.28} & \textbf{14.85}
    & \textbf{36.21} \\
\quad +Error LLM
    & {\textbf{67.00}} & {43.93}       & {25.23}          & {39.81} & {36.68}
    & 25.00                   & 22.43                       & 6.23                        & 11.24 & 9.74
    & 28.00 \\[2pt]
\multicolumn{12}{l}{\quad\textit{Creation}} \\
\quad +Error (ours)
    & {\textbf{49.00}} & {\textbf{40.37}} & {\textbf{23.37}} & {31.62} & {15.66}
    & \textbf{28.83}          & \textbf{23.23}              & \textbf{7.87}               & 22.20 & 23.58
    & 26.02 \\
\quad +Error LLM
    & {27.17}      & {27.73}          & {16.20}          & {\textbf{47.26}} & {\textbf{30.26}}
    & 19.83                   & 17.60                       & 4.70                        & \textbf{23.30} & \textbf{36.53}
    & \textbf{26.60} \\
\midrule
%% ── 35B Author ───────────────────────────────────────────────────────────────
\multicolumn{12}{l}{\textit{Skill Author: Qwen3.5-35B-A3B}} \\[2pt]
\multicolumn{12}{l}{\quad\textit{Deepening}} \\
\quad +Error (ours)
    & \textbf{65.00}          & \textbf{44.80}              & \textbf{25.17}              & 68.32 & 38.93
    & {27.00}      & {\textbf{22.20}} & {8.20}           & {\textbf{11.73}} & {\textbf{13.38}}
    & \textbf{32.58} \\
\quad +Error LLM
    & 37.83                   & 22.93                       & 12.83                       & \textbf{77.05} & \textbf{42.05}
    & {\textbf{30.50}} & {20.17}       & {\textbf{8.73}} & {9.95} & {12.73}
    & 28.80 \\[2pt]
\multicolumn{12}{l}{\quad\textit{Creation}} \\
\quad +Error (ours)
    & \textbf{27.17}          & \textbf{28.90}              & \textbf{18.53}              & \textbf{81.38} & \textbf{45.61}
    & {\textbf{24.00}} & {\textbf{21.00}} & {\textbf{6.53}} & {\textbf{32.80}} & {31.75}
    & \textbf{34.45} \\
\quad +Error LLM
    & 22.00                   & 27.67                       & 16.60                       & 54.61 & 37.93
    & {23.50}      & {16.87}          & {4.93}           & {11.24} & \textbf{33.75}
    & 26.49 \\
\bottomrule
\end{tabular}
}%

\end{table*}

\section{Selective Patch Aggregation Details}
\label{app:selective_patch_aggregation}

The main experiments apply all learned patches after hierarchical consolidation; selective aggregation instead asks whether some subset of patches yields a better skill.
We reuse the formalization of \cref{sec:method:skill}: $\mathcal{S}$ is a skill, $\pi_\theta$ the fixed agent, and $\mathcal{P}$ the pool of trajectory-level patches from Stage~2 (\cref{sec:method:swarm}), with $c(p)$ the number of source trajectories supporting patch $p\in\mathcal{P}$.
We write $\mathcal{S}\oplus\mathcal{P}'$ for the skill obtained by applying a patch subset $\mathcal{P}'\subseteq\mathcal{P}$ to $\mathcal{S}$ through the Stage~3 diff application (\cref{sec:method:consolidation}), and $\mathrm{acc}_V(\mathcal{S})\triangleq\mathcal{P}(\mathcal{S};\pi_\theta,V)$ for its validation pass rate on a 32-task validation set $V$.
Both selectors below search for a subset whose materialized skill maximizes $\mathrm{acc}_V$.

\myparagraph{Greedy top-1 selection.}
At iteration $t$, we evaluate the current skill $\mathcal{S}_t$, collect the newly proposed patches $\mathcal{P}_t$, and form a candidate set $\mathcal{C}_t$ of the five highest-coverage patches with $c(p)\ge5$.
For each $p\in\mathcal{C}_t$, we estimate its contribution by an add-one and remove-one intervention:
\[
\begin{aligned}
\Delta^+(p)&=\mathrm{acc}_V(\mathcal{S}_t\oplus p)-\mathrm{acc}_V(\mathcal{S}_t),\\
\Delta^-(p)&=\mathrm{acc}_V(\mathcal{S}_t\oplus\mathcal{P}_t)\\
&\quad -\mathrm{acc}_V\!\bigl(\mathcal{S}_t\oplus(\mathcal{P}_t\setminus\{p\})\bigr).
\end{aligned}
\]
Patches with $\Delta^+(p)<0$ or $\Delta^-(p)<0$ are discarded; the remainder are ranked by $r(p)=\Delta^+(p)+\Delta^-(p)$, with coverage as the tie-breaker, and the next skill is $\mathcal{S}_{t+1}=\mathcal{S}_t\oplus p_t^\star$ for the top-ranked patch $p_t^\star$.
\cref{sec:analysis:selective_patch_aggregation} reports the SpreadsheetBench-Vrf results.

\myparagraph{Bayesian optimization over patch subsets.}
BO fixes a patch universe $\mathcal{P}=\{p_j\}_{j=1}^m$ from the Stage-2 pool (\cref{sec:method:swarm}), keeping the top $m\le15$ patches by coverage.
It searches binary inclusion vectors $x\in\{0,1\}^m$, writing $\mathcal{P}_x=\{p_j:x_j=1\}$ for the selected subset and starting from the initial skill $\mathcal{S}_0$.
The objective is
\[
f(x)=\mathrm{acc}_V(\mathcal{S}_0\oplus\mathcal{P}_x)-\lambda\|x\|_0,
\qquad \lambda=0.
\]
The initial design evaluates the empty subset, all singletons, five random mixed subsets, and the full subset.
After each batch, evaluated subsets are split into a good set $G$ containing the top $\gamma=0.2$ fraction by $f(x)$ and a bad set $B$ containing the rest.
With Beta smoothing $\alpha=1$, the per-patch Bernoulli inclusion rates are
\[
\mu^G_j=\frac{\sum_{x\in G}x_j+\alpha}{|G|+2\alpha},\qquad
\mu^B_j=\frac{\sum_{x\in B}x_j+\alpha}{|B|+2\alpha}.
\]
Candidate subsets are sampled from $\phi_j=(1-\rho)\mu^G_j+\rho/2$ with $\rho=0.1$ and ranked by the TPE acquisition
\[
A(x)=\sum_{j=1}^{m}\bigl[\log q(x_j;\mu^G_j)-\log q(x_j;\mu^B_j)\bigr],
\qquad q(b;\mu)=\mu^b(1-\mu)^{1-b}.
\]
We evaluate the top eight candidates per round from a pool of 500, stop after four rounds or three rounds without improvement, and return the best observed subset.
\cref{sec:analysis:selective_patch_aggregation} compares the BO-selected +Error skill against applying all +Error patches.

\section{Broader Application Details}
\label{app:broader_application}

We extend \systemname{} to three broader document-agent settings: PDF extraction, PPTX editing, and DOCX editing.
All settings use Anthropic's official document skills as the underlying skill family \citep{anthropic2026documentskills}.
Source traces are used only for skill evolution, while held-out target tasks are used only for evaluation.
Each evaluation uses the task's exact local verifier and reports the same task-level pass-rate metric as the main experiments.
\cref{tab:broader_application} summarizes the transfer setting used for each modality in the main text; ``Base'' denotes the immediate skill before the listed source traces are applied.

\myparagraph{PDF extraction.}
The PDF study adapts VRDU and VAREX into local PDF-to-JSON tasks \citep{wang2023vrdu,barzelay2026varex}.
We use VRDU Registration Forms as the source corpus because it provides a coherent real collection of visually rich form-extraction traces with recurring field and layout conventions.
We evaluate on VAREX Flat as a separate held-out structured-extraction benchmark, testing whether those induced form-extraction procedures transfer across datasets rather than to more examples from the same corpus.
The reported PDF score uses the strongest validated VRDU Registration source split; other Registration splits are also positive but yield smaller gains.

\myparagraph{PPTX editing.}
The PPTX study uses TSBench presentation-editing tasks from Talk-to-Your-Slides \citep{jung2026talktoyourslides}.
The PPTX result evolves from TSBench training traces on one set of decks and evaluates on a deck-disjoint TSBench OOD split of held-out decks, testing transfer across presentation files rather than more edits from the same decks.

\myparagraph{DOCX editing.}
The DOCX study uses OfficeBench as the held-out target because it contains realistic office-automation tasks with verifiable Word/DOCX outputs \citep{wang2024officebench}.
We reserve all 64 convertible OfficeBench DOCX subtasks only for evaluation at \texttt{max\_turns=100}.
Because the remaining real, verifiable DOCX agent data is too small after reserving this target set, we use generated office-document traces as the source distribution.
These traces cover reusable operations such as editing templates, preserving document structure, appending content across files, and producing required sidecar files.

\begin{table*}[t]
\centering
\setlength{\tabcolsep}{4pt}
\resizebox{\textwidth}{!}{%
\begin{tabular}{lllccl}
\toprule
\textbf{Domain} & \textbf{Source traces} & \textbf{Held-out evaluation} & \textbf{Base} & \textbf{\systemname{}} & \textbf{Gain} \\
\midrule
PDF  & VRDU Registration Forms & VAREX Flat validation & 76.9\% & 85.3\% & +8.4 pp \\
PPTX & TSBench training traces & TSBench deck-disjoint OOD & 72.5\% & 88.8\% & +16.3 pp \\
DOCX & Generated document-operation tasks & OfficeBench DOCX holdout & 79.7\% & 87.5\% & +7.8 pp \\
\bottomrule
\end{tabular}
}%
\caption{Broader document-agent transfer. Each row evolves an official Anthropic document skill from source traces and evaluates on a held-out target domain. Base and \systemname{} are target pass rates (\%); Gain is their difference in percentage points (pp).}
\label{tab:broader_application}
\end{table*}

\section{Head-to-Head Comparison with Concurrent Skill-Evolution Systems}
\label{app:prior_systems}

This appendix is the head-to-head complement to the apples-to-apples comparisons in \cref{sec:analysis:core_design}: it pits \systemname{} against three full concurrent skill-evolution systems---XSkill \citep{jiang2026xskillcontinuallearningexperience}, EvoSkill \citep{alzubi2026evoskillautomatedskilldiscovery}, and SkillGen \citep{ma2026skillgenverifiedinferencetime}---rather than isolating one design choice at a time. We keep it out of the main text for three reasons.
\textbf{Attribution:} a whole-system score conflates a design \emph{idea} with its base model, harness, and engineering, so it cannot say which factor drove a difference; \cref{sec:analysis:core_design} instead varies a single design choice under a shared trace pool, model, and harness, which is what licenses our causal claims.
\textbf{Scope:} these systems pursue different goals and were built around specific (often proprietary) base models, bespoke harnesses, and different task domains, whereas we deliberately study open models that self-evolve; transplanting them onto one benchmark shows behavior \emph{in our setting}, not a ceiling on their design, so we read the numbers conservatively.

We reproduce these pipelines following their official codebase adapted to one shared protocol: the same open base model Qwen3.5-122B-A10B served with vLLM, evolution on the SpreadsheetBench-Verified 0:200 slice, and evaluation on the held-out 200:400 slice scored by official instance accuracy (our Vrf metric).
\cref{fig:prior_systems} reports the result against the No Skill floor, the Human-Written skill that several systems also start from, and \systemname{} Deepening +Error/+Combined.
\systemname{} attains the highest Vrf ($65.83$ +Error, $69.83$ +Combined), above all three systems; the per-system settings and the deviations from each native pipeline are detailed below.

\begin{figure}[t]
    \centering
    \begingroup
    \setlength{\columnwidth}{0.90\linewidth}%
    \begin{tikzpicture}
\begin{axis}[
    width=\columnwidth,
    height=0.66\columnwidth,
    ybar,
    bar width=13pt,
    xmin=0.3, xmax=8.7,
    ymin=0, ymax=82,
    ytick={0,20,40,60,80},
    ylabel={SpreadsheetBench-Vrf (\%)},
    xtick={1,2,3,4,5,6,7,8},
    xticklabels={
        No Skill,
        {Human-\\Written},
        XSkill,
        SkillGen,
        {EvoSkill\\(Claude Code)},
        {EvoSkill\\(React)},
        {\systemname{}\\+Error},
        {\systemname{}\\+Combined}
    },
    xticklabel style={font=\footnotesize, rotate=35, anchor=east, align=right, yshift=-1pt},
    yticklabel style={font=\footnotesize},
    ylabel style={font=\small},
    nodes near coords,
    nodes near coords style={font=\footnotesize, /pgf/number format/fixed, /pgf/number format/precision=1},
    every node near coord/.append style={anchor=south},
    grid=major,
    grid style={black!8},
    axis line style={black!45},
    area legend,
    legend style={
        at={(0.5,1.06)}, anchor=south, draw=none, fill=none,
        font=\footnotesize, legend columns=3,
        /tikz/every even column/.append style={column sep=6pt}
    },
]
% reference skills (No Skill floor, Human-Written shared init)
\addplot[ybar, bar shift=0pt, draw=black!55, fill=black!18]
    coordinates {(1,27.67) (2,48.33)};
\addlegendentry{Reference skills}
% concurrent skill-evolution systems (same model + benchmark)
\addplot[ybar, bar shift=0pt, draw=blue!55!black, fill=blue!22]
    coordinates {(3,23.0) (4,27.5) (5,33.5) (6,59.5)};
\addlegendentry{Concurrent systems}
% ours
\addplot[ybar, bar shift=0pt, draw=green!45!black, fill=green!55!black!50]
    coordinates {(7,65.83) (8,69.83)};
\addlegendentry{\systemname{} (ours)}
% Human-Written starting-skill reference line (drawn directly to avoid value labels)
\draw[dashed, thick, black!55] (axis cs:0.3,48.33) -- (axis cs:8.7,48.33);
\end{axis}
\end{tikzpicture}
    \endgroup
    \caption{Head-to-head comparison on SpreadsheetBench-Verified (Vrf, pass rate \%) with all systems using the same open base model (Qwen3.5-122B-A10B) and the held-out 200:400 test slice. Bars cover the No Skill floor, the Human-Written starting skill (dashed line), the three reproduced concurrent systems, and \systemname{} Deepening. EvoSkill's two bars are the same faithful reproduction differing only in agent harness (Claude Code, following original paper, vs.\ React, matching our setting).}
    \label{fig:prior_systems}
\end{figure}

\myparagraph{XSkill.}
XSkill is a non-parametric memory method that accumulates task-level procedure skills and action-level experiences, then retrieves, adapts, and injects them at inference; it was designed for multimodal tool-using agents.
We collect 200 trajectories on the 0:200 slice (same evolving set as \systemname{}), merge their per-trajectory skills into one 865-word global \texttt{SKILL.md} (batched LLM merge), and distill 20 curated action-level experiences.
At test time, for each instance we retrieve the top-3 experiences by lexical overlap, have the LLM adapt the global skill to the task, inject it into the system prompt, and run the agent for up to 100 turns.
This reproduction reaches $23.0$ Vrf. We replaced the lexical retriever with Qwen3-Embedding-0.6B. However, this achieves a slightly worse performance at $20.0$.

\myparagraph{EvoSkill.}
EvoSkill is a self-evolving loop---base agent, failure proposer, skill generator, validation evaluator, and a bounded frontier---designed for the Claude Code harness with Opus 4.5.
We mine failures on 0:160, validate candidates on 160:200, and evaluate the selected program on 200:400 (our SpreadsheetBench Vrf test subset), with a frontier of size 3; each iteration proposes and generates a replacement \texttt{xlsx/SKILL.md} and scores it on the validation slice, with Qwen3.5-122B-A10B served by vLLM.
We run two harnesses: a React-style local harness matching our setting reaches $59.5$ Vrf (5 iterations), while Claude Code accessed through an Anthropic-compatible proxy (closest to the native harness) reaches $33.5$ (8 workers, early-stopped near iteration 11).

\myparagraph{SkillGen.}
SkillGen induces a compact skill from contrastive failure/success analysis behind a paired verification gate before held-out evaluation.
We collect baseline trajectories on 0:200, cluster failure and success summaries with Qwen3-Embedding-0.6B ($k$-means) to synthesize contrastive observations, generate a compact \texttt{SKILL.md}, and verify it on 0:200 behind a net-gain acceptance gate (we use $\mathrm{min\_net\_gain\_abs}{=}1$; the upstream default is ${\ge}\,3$) for up to 8 refinement rounds, evaluating the accepted skill on 200:400.
From a $52.5$ baseline on the training slice, round 1 nets $0$ (rejected), round 2 nets $-4$ (rejected), and round 3 nets $+1$ (accepted), giving $27.5$ Vrf on the held-out slice. This improvement on training set but regression on test set might be attributed to the distribution shift: the training slice mixes 125 sheet-level and 75 cell-level tasks while the test slice is cell-only.

These head-to-head comparisons complement, and do not replace, the confounder-free apples-to-apples comparisons in \cref{sec:analysis:core_design}, on which our claims rest.

\section{Skill-Creator Baseline}
\label{app:skill_creator_baseline}

For the external Vrf-only baseline, we used Anthropic's official \texttt{skill-creator} skill for skill drafting and improvement \citep{anthropic2026skillcreatorrepo} through Claude Code with Opus 4.6 medium.
\cref{tab:skill_creator_vrf} reports SpreadsheetBench-Verified Vrf scores for the same skill-user/trace-source pairings used in the main spreadsheet evaluation, alongside the corresponding \cref{tab:main_v1} baselines and \systemname{} +Error results.
The resulting skills do not improve the corresponding base skill on average in either Deepening or Creation, while \systemname{} +Error remains consistently stronger.

\begin{table*}[t]
\centering
\setlength{\tabcolsep}{4pt}
\resizebox{\textwidth}{!}{%
\begin{tabular}{lccc ccc}
\toprule
\multirow{2}{*}{\textbf{User/Trace source}}
& \multicolumn{3}{c}{\textbf{Deepening}} & \multicolumn{3}{c}{\textbf{Creation}} \\
\cmidrule(lr){2-4}\cmidrule(lr){5-7}
& \textbf{Base} & \textbf{skill-creator} & \textbf{\systemname{} +Error}
& \textbf{Base} & \textbf{skill-creator} & \textbf{\systemname{} +Error} \\
\midrule
122B/122B & 48.33 & 27.33 & 65.83 & 26.17 & 26.67 & 49.00 \\
122B/35B  & 48.33 & 19.50 & 65.00 & 26.17 & 18.33 & 27.17 \\
35B/35B   &  9.67 & 18.50 & 27.00 & 20.17 & 17.67 & 24.00 \\
35B/122B  &  9.67 & 28.00 & 36.67 & 20.17 & 27.33 & 28.83 \\
\midrule
\textbf{Average} & \textbf{29.00} & \textbf{23.33} & \textbf{48.63} & \textbf{23.17} & \textbf{22.50} & \textbf{32.25} \\
\bottomrule
\end{tabular}
}%
\caption{SpreadsheetBench-Verified (Vrf) comparison with Anthropic's \texttt{skill-creator} baseline (pass rate, \%). For each skill-user/trace-source pair, \texttt{skill-creator} is compared with the corresponding Base skill and \systemname{} +Error under Deepening and Creation; the final row averages each column.}
\label{tab:skill_creator_vrf}
\end{table*}

The prompts used for this baseline were:

\begin{tcolorbox}[
    title={\textbf{Skill Creator Baseline Prompt: Deepening}},
    colback=gray!5, colframe=gray!50, fonttitle=\small\bfseries,
    breakable, left=4pt, right=4pt, top=4pt, bottom=4pt
]
\small\ttfamily\raggedright\noindent\detokenize{I have an agent running spreadsheet jobs using an xlsx skill. The agent's traces are provided in /path/to/traces/, where error and success traces are annotated with *_FAILURE.md and *_SUCCESS.md. Your job is to deepen the xlsx (/path/to/xlsx) to improve the agent's future performance when using the new skill. You should first induce the common and generalizable failure and success patterns from the traces and patch the xlsx skill using the skill-creator skill.}
\end{tcolorbox}

\begin{tcolorbox}[
    title={\textbf{Skill Creator Baseline Prompt: Creation}},
    colback=gray!5, colframe=gray!50, fonttitle=\small\bfseries,
    breakable, left=4pt, right=4pt, top=4pt, bottom=4pt
]
\small\ttfamily\raggedright\noindent\detokenize{I have an agent running spreadsheet jobs. The agent's traces are provided in /path/to/traces/, where error and success traces are annotated with *_FAILURE.md and *_SUCCESS.md. Your job is to create a spreadsheet skill to improve the agent's future performance equipped with the skill. You should first induce the common and generalizable failure and success patterns from the traces and then create the skill using the skill-creator skill.}
\end{tcolorbox}

\section{Prompt Templates and Intermediate Outputs}
\label{app:prompts}

\lstdefinestyle{appendixprompt}{
    basicstyle=\scriptsize\ttfamily,
    breakatwhitespace=false,
    breaklines=true,
    columns=fullflexible,
    keepspaces=true,
    showstringspaces=false,
    tabsize=2
}

This appendix reproduces the key prompt templates used in each pipeline stage and illustrates representative intermediate outputs to make the pipeline fully transparent and reproducible.

% ─────────────────────────────────────────────────────────
\subsection{Stage 1: Agent System Prompt Template}
\label{app:stage1_prompt}

The agent $\pi_\theta$ operates under the following system prompt during trajectory collection.
The skill $\mathcal{S}_0$ is prepended to the user context at inference time. Note that this differs from the standard skill-loading process, where the agent initially has access only to skill descriptions. We simplify this by preloading the SKILL.md content into the system prompt because \systemname{} focuses on improving a fixed target skill that is known relative to the task. Therefore, there is no need for the standard skill-selection step. Importantly, the \systemname{} skill-using agent still needs to procedurally discover resources referenced by the preloaded SKILL.md (e.g., resources and scripts), which are not preloaded.

\begin{tcolorbox}[
    title={\textbf{Stage 1 — Agent System Prompt (abbreviated)}},
    colback=gray!5, colframe=gray!50, fonttitle=\small\bfseries,
    breakable, left=4pt, right=4pt, top=4pt, bottom=4pt
]
\small
\textbf{Role:} You are an expert {role} (e.g., spreadsheet analysis) agent. \\[2pt]
\textbf{Skill context:} \texttt{[Contents of \(\mathcal{S}_0\) inserted here]} \\[2pt]
\textbf{Task:} Describing tasks and input files. \\[2pt]
\textbf{Tools available:} Describing tools and environment. E.g., \texttt{bash} (shell execution) for ReAct with file system access. \\[2pt]
\textbf{Format:} ReAct-style interaction --- alternate between reasoning traces
and tool calls until the task is complete.
\end{tcolorbox}

% ─────────────────────────────────────────────────────────
\subsection{Stage 2: Analyst Prompt Templates and Example Patches}
\label{app:stage2_prompts}

In Stage 2, the patch proposing agents first draw error and success memory items similar to \citep{ouyang2026reasoningbankscalingagentselfevolving}, which are generalizable trajectory-level knowledge that might be helpful for future task executions. Next, the agents read the original skill directory and then propose a patch to encode the memory items into the skill.

\subsubsection{Error Analyst Prompt ($\mathcal{A}^-$)}

\begin{tcolorbox}[
    title={\textbf{Error Analyst System Prompt (abbreviated)}},
    colback=red!3, colframe=red!30, fonttitle=\small\bfseries,
    breakable, left=4pt, right=4pt, top=4pt, bottom=4pt
]
\small
\textbf{Role:} You are an expert failure-analysis agent for \{domain\} tasks. \\[4pt]
\textbf{Mission:} Given an agent's execution artifacts (logs + produced files) and the
ground-truth solution, diagnose \emph{why the agent failed}, identify \emph{causal failure
reasons}, and \emph{validate} your diagnosis by implementing a minimal fix that makes the
agent output match the ground truth.
Your analysis must be \textbf{systematic}, \textbf{evidence-driven}, and \textbf{reproducible}.
\textbf{Do not guess} when you can verify. \\[4pt]
\textbf{Required Workflow (MANDATORY):} \\
1. Understand the task and failure surface --- identify exactly what is wrong in
   the output. \\
2. Trace the failure to agent behavior --- locate the decision or code step that produced
   the mismatch. \\
3. Validate the root cause with a minimal fix --- write fixed output and
   re-evaluate against the ground truth. \\
4. Re-evaluate --- if still failing, return to steps 1--3 and revise your diagnosis. \\[4pt]
\textbf{Output:} Produce (1) \emph{Failure Cause Items} --- systematic, causal reasons
grounded in observable agent behavior; (2) \emph{Failure Memory Items} ($\leq$3) ---
generalizable insights the agent should remember to avoid similar failures.
\end{tcolorbox}

\subsubsection{Success Analyst Prompt ($\mathcal{A}^+$)}

\begin{tcolorbox}[
    title={\textbf{Success Analyst System Prompt (abbreviated)}},
    colback=green!3, colframe=green!30, fonttitle=\small\bfseries,
    breakable, left=4pt, right=4pt, top=4pt, bottom=4pt
]
\small
\textbf{Role:} You are an expert in success pattern analysis for AI agent systems. \\[4pt]
\textbf{Mission:} Given a successful agent trajectory, identify \emph{generalizable}
behavior patterns that contributed to the correct answer. \\[4pt]
\textbf{Requirements:} \\
\textbf{Broad Coverage} --- every effective behavior in the trajectory must be captured by
a pattern. \\
\textbf{Frequency Awareness} --- patterns covering more instances should be listed first;
rare behaviors should be absorbed into the nearest broader pattern. \\
\textbf{Generalization} --- each pattern must describe a general mechanism, not a
single task-specific detail. \\[4pt]
\textbf{Output:} A compact set of \emph{Success Memory Items} with title, description,
and concrete examples of the effective behaviors observed.
\end{tcolorbox}

\subsubsection{Example Trajectory-Level Patch (Error Analyst Output)}
\label{app:example_trajectory_patch}

Each analyst produces a structured patch for a single trajectory.
The following example is from an error analyst ($\mathcal{A}^-$) processing one failure trajectory where the agent deleted rows outside the specified range.
The patch proposes adding a new \texttt{SKILL.md} section on row-manipulation safety; this patch, along with 52 others citing related structural-edit failures, is later consolidated into the ``Structural-edit safety'' SoP (\cref{sec:analysis:sops}).

\begin{tcolorbox}[
    title={\textbf{Example Stage 2 Trajectory-Level Patch (error analyst, one trajectory)}},
    colback=orange!3, colframe=orange!40, fonttitle=\small\bfseries,
    breakable, left=4pt, right=4pt, top=4pt, bottom=4pt
]
\small
\textit{Analyst reasoning:} Single error record reveals a clear failure pattern: the agent deletes rows outside the specified range and fails to distinguish partially-empty data rows from completely empty padding rows. The skill lacks any guidance on row/column manipulation or range boundary enforcement.

\medskip
\begin{lstlisting}[style=appendixprompt]
--- a/SKILL.md
+++ b/SKILL.md
@@ -55,0 +55,26 @@ ## Formula Verification Checklist

+## Row and Column Operations
+
+### Range Boundary Enforcement
+- When a task specifies a cell range (e.g., F14:M83), operations
+  MUST stay within those exact boundaries
+- Parse range format: start_col + start_row : end_col + end_row
+- Never apply operations to rows/columns outside the range
+
+### Row Deletion Logic
+Before deleting any row, verify BOTH conditions:
+1. Target columns are empty (as specified by task)
+2. Row falls within the explicit deletion range
+
+**Do NOT delete** completely empty rows across ALL columns unless
+explicitly instructed -- these often serve as structural padding.
+
+### Post-Operation Validation
+After row/column operations:
+- [ ] Count remaining rows matches expected count
+- [ ] No header/footer/padding rows outside range were affected
+- [ ] Data integrity is maintained in adjacent columns
+- [ ] Range boundaries were respected (no off-by-one errors)
\end{lstlisting}
\end{tcolorbox}

% ─────────────────────────────────────────────────────────
\subsection{Stage 3: Merge Operator Prompt and Example Consolidated Patch}
\label{app:stage3_prompts}

\subsubsection{Merge Operator Prompt ($\mathcal{M}$)}

\begin{tcolorbox}[
    title={\textbf{Merge Operator System Prompt (full)}},
    colback=blue!3, colframe=blue!30, fonttitle=\small\bfseries,
    breakable, left=4pt, right=4pt, top=4pt, bottom=4pt
]
\small
You are a skill edit coordinator. You receive multiple independently-proposed
patches that each suggest changes to a skill folder. Your job is to merge them
into one coherent, non-redundant patch. \\[4pt]
\textbf{Guidelines:} \\
1. \textbf{Deduplicate}: When multiple patches propose the same or very similar
   edits, keep the best version (most specific, best worded). \\
2. \textbf{Resolve conflicts}: If patches propose contradictory edits to the same
   section, choose the one with stronger justification or synthesize both into a
   better edit. \\
3. \textbf{Preserve unique insights}: Different patches address different
   failures --- include all unique, non-redundant edits. \\
4. \textbf{Maintain conciseness}: The merged patch should have $\leq$ the sum of
   unique edits across all input patches. Remove redundancy. \\
5. \textbf{Ensure independence}: Edits in the merged patch MUST be line-level
   independent --- no two edits may target overlapping lines or the same passage
   of text, even across different operations. \\
6. \textbf{Atomic create/link pairs}: A \texttt{create} operation for
   \texttt{references/*.md} and the \texttt{SKILL.md} edit that inserts a link to
   it are an inseparable pair --- keep both or drop both. \\[4pt]
\textbf{Prevalent pattern bias:} When multiple patches independently propose
similar edits addressing the same class of failure or success pattern, treat
this recurrence as evidence of a \emph{systematic} property of the task.
Preserve such prevalent edits with higher priority and express them as
general principles rather than instance-specific fixes.
\end{tcolorbox}

\subsubsection{Example Final Consolidated Patch $p^*$ (After Full Merge Hierarchy)}
\label{app:example_final_patch}

The following excerpt shows the reasoning and representative edits from the final consolidated patch $p^*$ produced after four levels of hierarchical merging over 323 individual trajectory patches on SpreadsheetBench-Verified.

\begin{tcolorbox}[
    title={\textbf{Example Stage 3 Final Patch Output (excerpt)}},
    colback=purple!3, colframe=purple!30, fonttitle=\small\bfseries,
    breakable, left=4pt, right=4pt, top=4pt, bottom=4pt
]
\small
\begin{lstlisting}[style=appendixprompt]
{
  "reasoning": "Merged 3 patches addressing mixed failure/success
    evidence. Key consolidation decisions: (1) Synthesized recalc.py
    workflow from all patches using the most prominent CRITICAL WARNING
    placement, CSV fallback validation, and a verification loop;
    (2) Consolidated library selection guidance into a comprehensive
    decision tree; (3) Combined row deletion guidance emphasizing
    bottom-up/right-to-left deletion order from all patches;
    (4) Merged formula validation checklists without redundancy.",
  "edits": [
    {
      "file": "SKILL.md",
      "op": "insert_after",
      "target_section": "# Requirements for Outputs",
      "content": "## Important Automation Guidelines\n\n
        **Prefer Python over VBA for Automation**: When tasks request
        VBA macros or spreadsheet automation, implement the logic in
        Python using openpyxl/pandas instead. This provides better
        error handling, easier debugging, cross-platform compatibility,
        and avoids macro security issues."
    },
    {
      "file": "SKILL.md",
      "op": "insert_after",
      "target_section": "## Important Requirements",
      "content": "### When to Compute Directly in Python\n
        If LibreOffice/recalc.py cannot evaluate certain formula
        patterns (complex array formulas, advanced INDEX/MATCH
        combinations), compute the logic directly in Python and write
        results as values. Use this fallback when formula returns errors
        after recalculation despite correct syntax."
    }
  ],
  "changelog_entries": [
    "Added automation guidelines preferring Python over VBA",
    "Consolidated recalc.py warnings into a single CRITICAL WARNING"
  ]
}
\end{lstlisting}
\end{tcolorbox}

\subsubsection{Example Translated Diff: Final Consolidated Patch $p^*$}
\label{app:example_patch}

The JSON patch $p^*$ produced by Stage~3 is translated programmatically into a unified diff and applied directly to the skill directory.
The following excerpt shows two hunks from the final applied diff on SpreadsheetBench-Verified, corresponding to the two most prevalent SoPs: formula recalculation enforcement (55.1\% of patches) and tool-selection guidance (54.8\% of patches).

\begin{tcolorbox}[
    title={\textbf{Example Stage 3 Applied Diff (excerpt from final $p^*$)}},
    colback=purple!3, colframe=purple!30, fonttitle=\small\bfseries,
    breakable, left=4pt, right=4pt, top=4pt, bottom=4pt
]
\small
\begin{lstlisting}[style=appendixprompt]
--- a/SKILL.md
+++ b/SKILL.md
@@ -126,3 +261,18 @@ ## Common Workflow

 ## Common Workflow

+### CRITICAL WARNING: Formula Recalculation Is Mandatory
+
+**If you write ANY formulas to an Excel file using openpyxl,
+you MUST run recalc.py before considering the task complete.**
+
+Formulas written via openpyxl exist only as text strings until
+recalculated. Without running recalc.py:
+- Cells return None/empty when read with data_only=True
+- Evaluation fails even if formulas are syntactically correct
+- The output file is incomplete
+
+This is non-negotiable. Do not proceed to verification or
+delivery until recalc.py confirms success.
+
@@ -138,3 +285,9 @@ ### Working with openpyxl

+### Tool Selection Warning
+
+**CRITICAL**: When modifying spreadsheets that contain existing
+formulas you need to preserve:
+- Use openpyxl (load_workbook() then save()) -- formulas remain
+  as strings
+- Avoid pandas (to_excel()) -- silently converts formulas to
+  static values permanently
\end{lstlisting}
\end{tcolorbox}

\end{document}